\documentclass{article} % For LaTeX2e
\usepackage{iclr2026_conference,times}

\usepackage{amsmath,amsfonts,bm}

\def\eqref#1{equation~\ref{#1}}
\def\1{\bm{1}}

\DeclareMathAlphabet{\mathsfit}{\encodingdefault}{\sfdefault}{m}{sl}
\SetMathAlphabet{\mathsfit}{bold}{\encodingdefault}{\sfdefault}{bx}{n}

\usepackage{hyperref}
\usepackage{url}
\usepackage{booktabs}
\usepackage[table]{xcolor}
\usepackage{algorithm}
\usepackage{algpseudocode}
\usepackage{graphicx}
\usepackage{wrapfig}

\title{RL-Native Distillation: Exploiting Scored Trajectories for Few-Step Image Generation}

\author{
\hspace{1em}Yuhan Li$^{1,*}$  \hspace{0.8em}
Fangao Zeng$^{2,*,\dagger}$  \hspace{0.8em}
Sicong Kang$^{2,*}$  \hspace{0.8em}
Mengfei Xu$^{2}$   \hspace{0.8em}
Hao Zhou$^{2}$  \vspace{-2mm} \AND\vspace{2mm} 
\hspace{1em}Wenxiang Shang$^{2}$ \hspace{1em}
Wei Li$^{2}$ \hspace{1em}
Pipei Huang$^{2}$ \hspace{1em}
Bo Zheng$^{2, \ddagger}$ \hspace{1em}
Bingbing Ni$^{1, \ddagger}$ \\
\vspace{1mm}
\hspace{12mm}
$^*$Equal Contribution \quad
$^\dagger$Project Leader \quad
$^\ddagger$Joint Corresponding Authors
\\
\hspace{13mm}
$^1$Shanghai Jiao Tong University \quad
$^2$Taobao \& Tmall group of Alibaba \vspace{1mm}
\vspace{-6mm}
}

 \iclrfinalcopy % Uncomment for camera-ready version, but NOT for submission.
\begin{document}

\maketitle

\begin{abstract}
Efficient text-to-image generation requires both reinforcement-learning (RL)-based reward alignment and few-step distillation, yet these procedures are typically performed sequentially, increasing training cost and risking the loss of reward gains during compression. We take an \emph{RL-native} perspective: diffusion RL already generates reward-scored finite-step trajectories, whose intermediate states provide distillation supervision. Based on this insight, we propose \textbf{REST} (Reward-Enhanced Scored-Trajectory Distillation), a single-stage co-training framework in which a decoupled student learns from the evolving RL teacher's trajectories without changing teacher optimization. Advantage-Modulated Distillation (AMD) transforms rollout advantages into signed weights, strengthening imitation of preferred trajectories and aligning distillation priorities with task value. The resulting framework is general and lightweight, requires no extra image rollouts, no separate distillation dataset, and no adversarial training. Experiments on compositional generation, visual text rendering, and human-preference alignment demonstrate competitive few-step, CFG-free generation with RAM or DiffusionNFT teachers. With only four sampling steps, REST-RAM achieves a DrawBench PickScore of 23.97, outperforming both the 40-step RAM teacher (23.95) and RTDMD (23.71).
\end{abstract}   
\section{Introduction}
\label{sec:intro}

Recent progress in diffusion and flow-matching models~\citep{flux, wu2025qwenimagetechnicalreport} has substantially improved the visual quality of text-to-image generation. In real practice, however, two post-training procedures are often required before such models become truly useful~\citep{dmdr, fan2026rdm}: few-step sampling and classifier-free guidance (CFG)~\citep{ho2022cfg} distillation~\citep{lcm, dmd} for efficient inference, and reinforcement-learning-based alignment for human preference~\citep{shao2024grpo}.  These two procedures are traditionally executed sequentially, which is cumbersome and often suboptimal: distillation may wash out the reward gains obtained from RL, while subsequent RL can disturb a previously distilled few-step generator and degrade its structure or few-step inference stability~\citep{dmdr}.

\begin{wrapfigure}{rt}{0.6\textwidth}
  \centering
  \includegraphics[width=\linewidth]{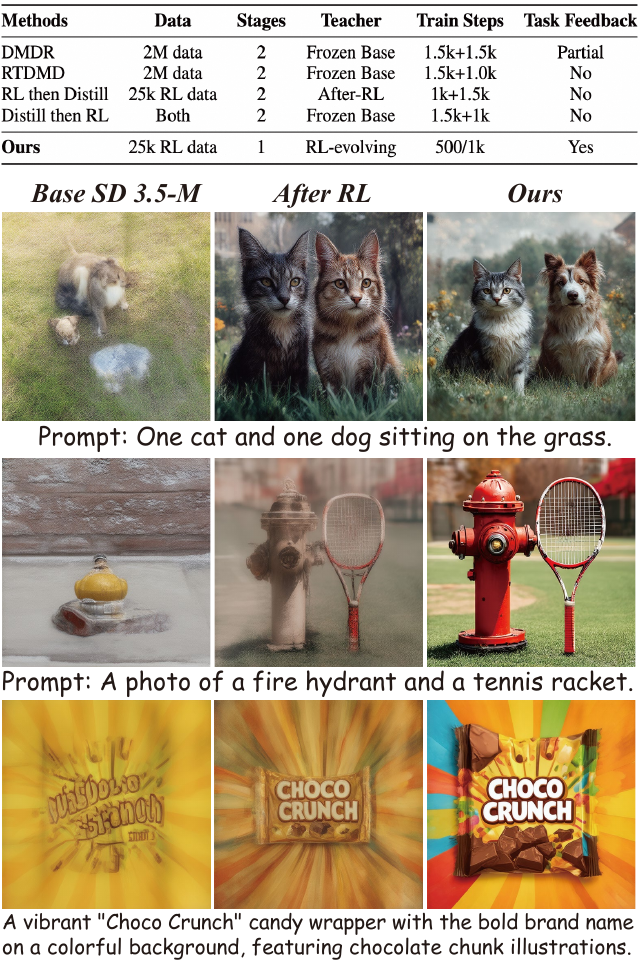}
  \vspace{-6mm}
  \caption{REST achieves few-step, reward-aligned generation with simple, efficient training (top). Few-step CFG-free samples on PickScore, OCR, and GenEval (bottom) show that REST improves on RL-trained models' few-step capabilities, yielding higher quality and preference alignment.}
  \label{fig:motivation}
\end{wrapfigure}

A natural question is whether reward alignment and few-step distillation can be performed in a unified post-training stage. As shown in Fig.~\ref{fig:motivation}, recent works such as DMDR~\citep{dmdr} and RTDMD~\citep{rtdmd} make progress in this direction by introducing a warm-up distribution-matching distillation (DMD)~\citep{dmd} before RL, followed by RL+DMD parallel optimization. The multi-stage system often involves balancing stage-specific hyperparameters and using more data and training iterations. They also suffer from complicated pipelines compared to RL algorithms, such as extra rollouts, and parallel training. More importantly, the frozen RL-agnostic teacher distribution used by the distillation objective may pull against the continuously evolving reward-optimized policy, creating an inherent tension between imitation and reward improvement.

In this work, we take an \emph{RL-native} perspective: distillation can be built directly into RL training by reusing its reward-scored rollouts and intermediate states, without a separate pre-distillation stage or external distillation data. We observe that RL training already improves direct few-step generation (Fig.~\ref{fig:motivation}), but its reward objective does not explicitly train the model to preserve full-step generation quality under a shorter compressed sampling. To learn this compression while preserving the original RL optimization, we attach a decoupled student branch that performs segment-wise imitation along the teacher's existing reward-scored trajectories. The teacher continues to improve through RL, while the student learns a few-step, CFG-free generator from the evolving teacher.

Few-step distillation is inherently lossy, leading to structural ambiguity and fine-grained artifacts, and we aim to minimize the loss of semantically important content. Inspired by~\citet{awm}, we further reuse the rollout reward in student distillation through \emph{Advantage-Modulated Distillation} (AMD). AMD applies an affine transformation to the reward advantage of each teacher rollout, yielding a signed modulation coefficient for the base distillation loss. AMD incorporates reward feedback to encourage the student to preserve task-relevant generation capabilities during compression. This mechanism induces a contrastive effect analogous to CFG~\citep{ho2022cfg} and NFT~\citep{chen2026nft}: high-reward trajectories exert stronger attractive supervision toward desirable teacher behaviors, whereas sufficiently low-reward trajectories provide a mild repulsive gradient that discourages the student from reproducing undesirable ones. Compared with uniform distillation, this signed reward modulation improves both visual clarity and reward alignment. Because AMD only reweights the per-sample objective without changing its target, it also serves as a general reward-aware wrapper for different distillation losses.

In summary, we present \textbf{REST} (Reward-Enhanced Scored Trajectory Distillation), a unified RL-distillation co-training framework for efficient and reward-aligned diffusion generation. To our knowledge, it is the first single-stage framework for RL-distillation collaboration. First, it delivers high-quality few-step generation, matching or even surpassing the full-step inference RL-trained teacher: the general AMD reward modulation compensates for the quality loss introduced by pure distillation. Second, it is general: the decoupled teacher--student design and the generic AMD formulation make the method independent of a specific teacher RL algorithm or a particular distillation loss, as shown in Secs.~\ref{subsec:main_results} and~\ref{subsec:exp_analysis}. Third, it is simple and efficient: the framework avoids warm-up scheduling and delicate coordination among multiple competing objectives; it also requires no extra image rollouts and no separate distillation dataset as in DMDR and RTDMD.
\begin{itemize}
    \item We propose REST, a decoupled teacher--student co-training framework that attaches a lightweight student branch to an arbitrary RL teacher branch. The student reuses the teacher's existing rollout trajectory and reward scores, introducing no extra sampling cost while avoiding interference with teacher RL optimization.
    \item We introduce AMD, a general reward-aware modulation mechanism for arbitrary trajectory distillation losses. The scale-and-shift transformation of advantages helps align distillation priorities with task value, improving the quality of the few-step student.
    \item We conduct extensive experiments with RAM and DiffusionNFT, demonstrating competitive 4/8 step CFG-free few-step inference that approaches the quality of the full-inference RL teacher. Compared with two-stage pipelines, it requires much fewer training iterations and obtains substantially better alignment performance.
\end{itemize}

% This helps align distillation priorities with task value 
\section{Related Work}
\label{sec:related}

\noindent \textbf{Reinforcement learning for diffusion alignment.}
Reinforcement learning has become a major paradigm for aligning diffusion and flow-matching models with human preferences. Policy-gradient methods such as Flow-GRPO~\citep{liu2025flowgrpo}, DanceGRPO~\citep{xue2025dancegrpo}, MixGRPO~\citep{li2025mixgrpo}, AWM~\citep{awm} and DiffusionNFT~\citep{diffusionnft} optimize terminal rewards over denoising trajectories, while preference-based methods~\citep{dpo,diffusiondpo,spo} avoid explicit reward modeling through pairwise supervision. Direct-gradient approaches such as ReFL~\citep{xu2023imagereward}, DRaFT~\citep{draft}, DRTune~\citep{drtune}, and LeapAlign~\citep{leapalign} propagate reward gradients through differentiable sampling paths, which require differentiable rewards. Despite their success in improving reward-aligned sampling, these methods largely overlook the potential of integrating distillation into RL, and thus the possibility of achieving reward-aligned few-step generation remains underexplored.

\noindent \textbf{Few-step diffusion distillation.}
Few-step distillation aims to compress multi-step diffusion samplers into generators requiring only one or a few inference steps. Early trajectory-distillation methods directly regress the student toward the outputs of the teacher's ODE trajectory: Knowledge Distillation~\citep {luhman2021knowledge} matches the teacher's full sampling result in a single step, and Progressive Distillation~\citep{progressive_distil} iteratively halves the number of sampling steps by training the student to match two teacher steps at a time. While consistency-based methods, including Consistency Models~\citep{cm}, Latent Consistency Models~\citep{lcm}, and Phased Consistency Models~\citep{pcm}, enforce trajectory-level consistency, score-based methods such as SwiftBrush~\citep{swiftbrush} and Score Implicit Matching~\citep{sim} pursue fast generation through score distillation. Distribution Matching Distillation (DMD)~\citep{dmd} and DMD2~\citep{dmd2} improve few-step quality by matching student and teacher distributions with auxiliary fake-score or discriminator-like networks. However, these distillation objectives are usually reward-agnostic and often require a frozen teacher, a separate distillation stage, or additional auxiliary models. REST is orthogonal to the choice of base distillation loss: it modulates common distillation losses with reward-derived advantages, making distillation preference-aware without redesigning the underlying distillation algorithm.

\noindent \textbf{Unified RL-distillation training.}
Sequentially applying RL alignment and few-step distillation is cumbersome and can cause objective interference: distillation may wash out RL gains, while later RL can destabilize a previously distilled few-step generator. Recent works such as DMDR~\citep{dmdr} and RTDMD~\citep{rtdmd} integrate DMD-style distillation with RL, but their fake-score-centered designs involve multi-network training, warm-up, and staged optimization, and their frozen reference distributions may lag behind the continuously evolving reward-optimized policy. REST instead uses a decoupled dual-branch design: the teacher continues its original RL optimization, while the student simultaneously learns a few-step policy from the same reward-scored teacher rollouts through REST. This yields a unified RL-distillation co-training framework that requires no extra image rollouts, no separate distillation dataset, and no multi-stage model training.
  
\begin{figure}[t]
  \centering
  \includegraphics[width=1\linewidth]{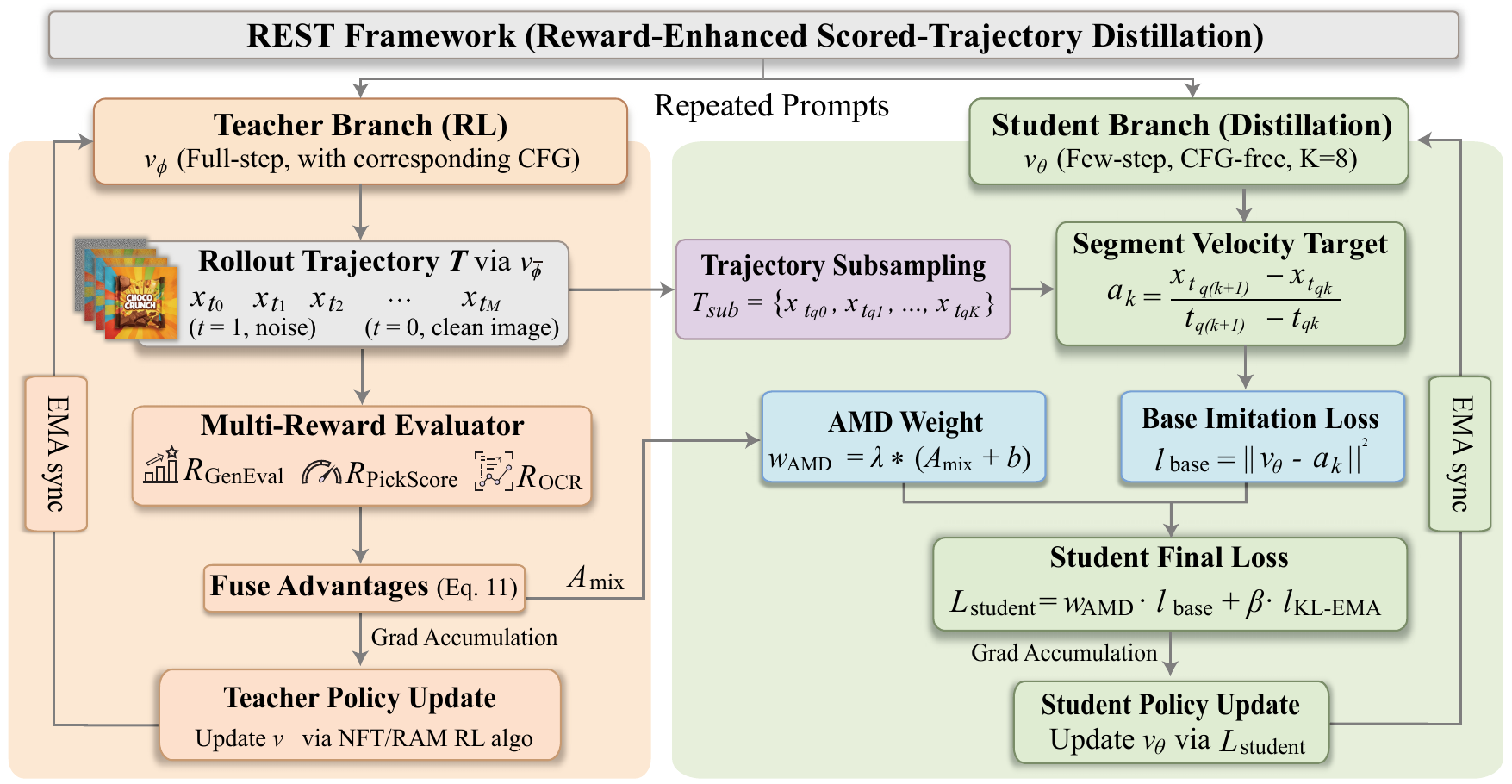}
  \caption{Overview of REST. The decoupled student reuses reward-scored teacher rollouts, while AMD modulates trajectory distillation without altering teacher RL optimization.}
  \label{fig:framework}
\end{figure}

\section{Method}
\label{sec:method}

\subsection{Preliminary: Forward-Process Diffusion RL}
\label{subsec:preliminary}

Our teacher branch can in principle be optimized by any diffusion RL algorithm, but it pairs especially well with forward-process RL algorithms. The representative instances include DiffusionNFT~\citep{diffusionnft}, AWM~\citep{awm} and RAM~\citep{bergmeister2026ram}, which all cast reward alignment as a reward-weighted velocity regression on the forward noising process $x_t=(1-t)x_0+t\epsilon$, where $x_0$ is the rollout sample and $\epsilon\sim\mathcal{N}(0,I)$. DiffusionNFT converts a scalar reward into an optimality probability $r\in[0,1]$ and implicitly parameterizes a positive and a negative policy around the old velocity $v^{\rm old}$,
\begin{equation}
v_\phi^+ = (1-\beta)v^{\rm old}+\beta v_\phi(x_t), \qquad
v_\phi^- = (1+\beta)v^{\rm old}-\beta v_\phi(x_t),
\end{equation}
and trains the policy $v_\phi$ with
\begin{equation}
\mathcal{L}_{\rm NFT}(\phi) = \mathbb{E}\left[r\left\|v_\phi^+(x_t)-v^{gt}\right\|^2 + (1-r)\left\|v_\phi^-(x_t)-v^{gt}\right\|^2\right],
\label{eq:nft_loss}
\end{equation}
where $v^{gt}=\epsilon-x_0$ is the forward flow-matching~\citep{lipman2022flow} target. RAM instead directly regresses $v_\phi$ toward a reward-shifted version of the reference velocity,
\begin{equation}
\mathcal{L}_{\rm RAM}(\phi) = \mathbb{E}_t\left[\left\|v_\phi(x_t)-\operatorname{sg}\!\left(v^{\rm base}(x_t)+r(x_0)\big(v^{gt}-v^{\rm old}(x_t)\big)\right)\right\|^2\right],
\label{eq:ram_loss}
\end{equation}
where $\operatorname{sg}(\cdot)$ is the stop-gradient operator, $v^{\rm base}(\cdot)$ denotes the frozen base model, and $v^{\rm old}(\cdot)$ denotes the lagged policy used for sampling. In our framework, the teacher branch is optimized by any forward-process algorithm, e.g., Eq.~\ref{eq:nft_loss} or Eq.~\ref{eq:ram_loss}, producing an evolving teacher policy $v_\phi$ together with reward-scored rollouts that the student branch reuses.

\subsection{Decoupled Teacher--Student Co-training}
\label{subsec:teacher_student}

Given the teacher policy $v_\phi$ in Sec.~\ref{subsec:preliminary}, REST attaches a decoupled student branch $v_\theta$ that learns a few-step, CFG-free generator from the teacher's rollouts, without altering the teacher optimization, as shown in Fig.~\ref{fig:framework}. For each prompt $c$, the teacher branch samples with its ODE solver~\citep{song2020ddim} on an $M$-step schedule (e.g., $M=20$) with classifier-free guidance (CFG), producing a trajectory
\begin{equation}
\mathcal{T} = \left\{x_{t_0}, x_{t_1}, \ldots, x_{t_M}\right\}, \qquad 1=t_0>t_1>\cdots>t_M=0,
\end{equation}
and a terminal reward $r=R(x_0,c)$. The student, in contrast, is designed to run with only $K\ll M$ steps (e.g., $K=4\text{ or }8$) at inference time. Rather than sampling a separate trajectory for the student, we select a $K$-step subset of the teacher's schedule and re-index it as an imitation trajectory
\begin{equation}
\mathcal{T}_{\rm sub} = \left\{x_{t_{q0}}, x_{t_{q1}}, \ldots, x_{t_{qK}}\right\} \subset \mathcal{T}, \qquad \{t_{q0},\ldots,t_{qK}\}\subset\{t_0,\ldots,t_M\},
\end{equation}
so that the student can be trained directly on the teacher's existing rollout, with no additional sampling required. For each student step $k$, the \emph{piecewise trajectory velocity} of the corresponding segment is directly computed from the teacher's rollout,
\begin{equation}
v^{\rm gt}_k = \frac{x_{t_{q(k+1)}}-x_{t_{qk}}}{t_{q(k+1)}-t_{qk}}.
\label{eq:segment_velocity}
\end{equation}
In this decoupled co-training framework, the student model $\theta$ is trained to imitate this piecewise trajectory, as detailed in Sec.~\ref{subsec:awr}, thereby inheriting the teacher's generation quality while enabling few-step CFG-free sampling.

This dual-branch design brings two benefits. First, it is a one-stage, end-to-end training framework: the student is trained by reusing the teacher's existing rollout trajectory, making training more efficient than a two-stage pipeline. Second, the teacher policy $v_\phi$ continues to evolve under RL training, avoiding a frozen teacher that could hold back the student branch's RL progress.

\subsection{Advantage-Weighted Regression for Teacher Tracking}
\label{subsec:awr}

Advantage-weighted regression~\citep{peters2007rwr, peng2019awr, kostrikov2021iql}, \emph{i.e.} AWR, recasts policy improvement as a reward-weighted supervised regression problem: it treats each observed state-action pair as a demonstration and re-weights its log-likelihood under the current policy by the advantage of that action, so that high-advantage demonstrations receive greater weight. We adopt this view to let the student track the teacher's reward-scored rollout: each teacher trajectory segment is treated as a state-action demonstration, whose imitation strength is modulated by the reward of the rollout it comes from. For prompt condition $c$, the state is $s_{k}=\left(x_{t_{qk}}, t_{qk}, c\right)$ and the demonstrated action $a_{k}=v^{{\rm gt}}_k$ is given by the teacher's segment velocity (Eq.~\ref{eq:segment_velocity}). We interpret the student’s velocity prediction as the mean of an implicit, fixed-variance Gaussian policy following \citet{liu2025flowgrpo} over teacher segment velocities,
\begin{equation}
\pi_\theta(a_k\mid s_k) = \mathcal{N}\!\left(a_k;\, v_\theta(s_k), \sigma^2 I\right),
\label{eq:gaussian_policy}
\end{equation}
under which the negative log-likelihood of the demonstrated action is, up to an additive constant independent of $\theta$,
\begin{equation}
-\log \pi_\theta(a_k\mid s_k) = \frac{1}{2\sigma^2}\|a_k -v_\theta(s_k)\|^2 + \mathrm{const.}
\label{eq:gaussian_nll}
\end{equation}
Dropping the constant, this yields the plain \emph{per-step imitation loss}
\begin{equation}
\ell_{\rm base}(\theta;k) = \left\|v_\theta(s_{k}) - a_{k}\right\|^2 \;\propto\; -\log \pi_\theta(a_{k}\mid s_{k}),
\label{eq:base_loss}
\end{equation}
which treats every teacher trajectory segment as equally reliable supervision. AWR improves a policy by imitating demonstrated actions with weights given by their advantages,
\begin{equation}
\max_\theta \; \mathbb{E}_{(s,a)\sim\mathcal{D}}\left[A(s,a)\log\pi_\theta(a\mid s)\right].
\label{eq:awr_objective}
\end{equation}
Using Eq.~\ref{eq:gaussian_nll}, this objective induces an advantage-weighted regression loss over teacher trajectory segments, where the log-likelihood term is replaced by the base imitation loss $\ell_{\rm base}$.
Here, $\mathcal{D}$ is the reward-scored teacher trajectory dataset formed by all segments $(s_{k},a_{k})$. For each reward source $i$, its advantage $A^{(i)}(s_k,a_k)\in[-1,1]$ is the clipped group-normalized advantage of the rollout associated with prompt $c$, and is shared by all segments from the same rollout, as in Flow-GRPO~\citep{liu2025flowgrpo} and AWM~\citep{awm}. We further fuse different rewards on the normalized advantages level,
\begin{equation}
A_{\rm mix}=\frac{\sum_{i=1}^{N_R}\alpha_i A^{(i)}}{\sum_{i=1}^{N_R}\alpha_i}, \qquad \alpha_i\ge 0,
\label{eq:multi_reward_adv}
\end{equation}
where $N_R$ is the number of reward sources and $\alpha_i$ controls their contribution. We then introduce a global scale $\lambda = 1$ and a positive shift $b = 0.5$ to obtain a signed modulation coefficient, leading to our Advantage-Modulated Distillation (AMD) objective for teacher trajectory tracking:
\begin{equation}
\mathcal{L}_{\rm AMD}(\theta) = \mathbb{E}_{(s_k,a_k)\sim\mathcal{D}}\left[\lambda\left( A_{\rm mix}+b\right)\,\ell_{\rm base}(\theta;k)\right].
\label{eq:amd_loss}
\end{equation}
In practice, we further regularize the student against an EMA copy of itself, denoted by $v_{\theta_{\rm ema}}$, using a fixed-variance Gaussian KL surrogate
\begin{equation}
\mathcal{L}_{\rm KL\mbox{-}EMA}(\theta)=\mathbb{E}_{k}\left[\left\|v_\theta(s_k)-v_{\theta_{ema}}(s_k)\right\|^2\right],
\label{eq:kl_EMA}
\end{equation}
which stabilizes optimization and reduces the training jitter caused by the continuously changing teacher policy. The final student objective is therefore
\begin{equation}
\mathcal{L}_{\rm student}(\theta)=\mathcal{L}_{\rm AMD}(\theta)+\beta \mathcal{L}_{\rm KL\mbox{-}EMA}(\theta),
\label{eq:student_loss}
\end{equation}
where $\beta$ controls the strength of the EMA-student regularization.

\subsection{Understanding Advantage-Modulated Distillation}
\label{subsec:understanding_amd}

\noindent \textbf{Decomposing the AMD objective.} Equation~\ref{eq:amd_loss} can be decomposed into two additive terms,
\begin{equation}
\mathcal{L}_{\rm AMD}(\theta) = \underbrace{\lambda b\,\mathbb{E}_{(s_k,a_k)\sim\mathcal{D}}\!\left[\ell_{\rm base}(\theta;k)\right]}_{\text{imitation prior}} + \underbrace{\lambda\,\mathbb{E}_{(s_k,a_k)\sim\mathcal{D}}\!\left[A_{\rm mix}\,\ell_{\rm base}(\theta;k)\right]}_{\text{reward-driven correction}}.
\label{eq:amd_decomp}
\end{equation}
The first term is a constant-weighted imitation prior. A positive shift $b$ is crucial in our setting: the initial student branch lacks reliable image-generation capability, and needs a stable positive imitation signal to bootstrap its few-step generator. From another perspective, even trajectories with relatively low advantages still contain useful teacher dynamics for the student, especially at early training stages, as shown in Fig.~\ref{fig:ab_shift} in Sec.~\ref{subsec:ablation}. The second term is the reward-driven correction inherited from reward-weighted supervised regression: it biases the student toward teacher segments with high combined reward advantages, allowing the few-step student to concentrate on the most task-level important contents of the teacher's scored rollouts and potentially surpass the average behavior of the RL-trained full-step teacher. This reward-aware distillation effect is unavailable to conventional distillation methods~\citep{progressive_distil} that imitate teacher rollouts uniformly.

\noindent \textbf{AMD as a generic reward-aware distillation wrapper.} The segment-velocity MSE in Eq.~\ref{eq:base_loss} is a simple trajectory distillation objective~\citep{progressive_distil}, rather than a required design choice. Since AMD only changes the per-sample weight, it can also modulate other losses evaluated on the same reward-scored trajectories. We validate this flexibility with PCM-style phase consistency~\citep{pcm} in Sec.~\ref{subsec:exp_analysis}, without changing the teacher's RL objective.

\noindent \textbf{Aligning distillation priorities with task value.}
Few-step distillation introduces approximation errors, but their numerical magnitude does not necessarily reflect their impact on generation quality. For example, a small stroke that turns an ``O'' into a ``Q'' can make the requested text incorrect, whereas a comparable change in background texture may have little effect on task quality. AMD incorporates terminal reward feedback to prioritize imitation of high-quality teacher trajectories, allowing task value to influence distillation beyond uniform regression. This helps align distillation priorities with task value, consistent with the gains from advantage modulation observed in Sec.~\ref{subsec:ablation}.

\noindent \textbf{Auxiliary DMD for four-step generation.}
We find that DMD helps improve image sharpness and explore it as an auxiliary loss to complement AMD in four-step REST-RAM. For each sample, we uniformly select one of four stored teacher states aligned to the student's target input times $\{1.0,0.9,0.75,0.5\}$. From the selected state $x_t$, the student predicts a clean latent in one step:$\widehat{x}_0=x_t-tv_\theta(x_t,t,c)$.After re-noising this prediction, the evolving EMA teacher and a learned fake-score model provide the DMD update direction. We set the DMD-to-AMD loss weight ratio to $2:1$.

\begin{figure}[t]
  \centering
  \includegraphics[width=1\linewidth]{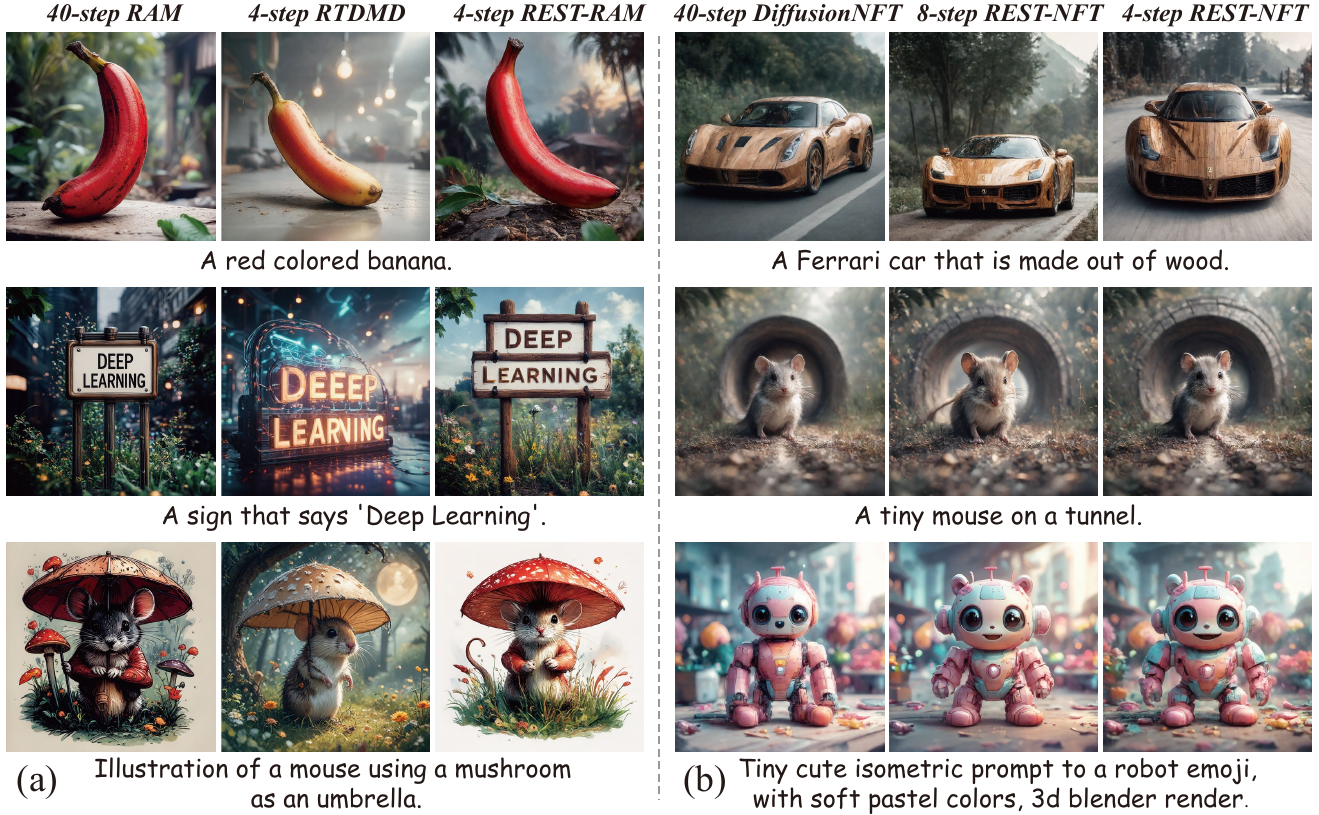}
  \vspace{-5mm}
  \caption{Four-step generation and compatibility with different RL teachers. (a) 4-step REST-RAM achieves visual quality comparable to the 40-step RAM teacher, with better structural coherence and prompt adherence than 4-step RTDMD. (b) 4-step and 8-step REST-NFT produce samples of comparable visual quality to the 40-step DiffusionNFT teacher.}
  \label{fig:4step}
\end{figure}

\noindent \textbf{Compatibility with different RL teachers.}
REST accesses the teacher through complete, time-indexed rollout trajectories and their terminal rewards, rather than through a particular policy-optimization objective. A diffusion RL method that exposes this interface can therefore provide supervision for REST. Our experiments with both RAM and DiffusionNFT demonstrate this compatibility at four and eight inference steps (Tab.~\ref{tab:sd35_results} and Fig.~\ref{fig:4step}). Appendix~\ref{app:nft_ram_similarity} further analyzes their shared velocity-regression structure.

\section{Experiments}
\label{sec:experiments}

\subsection{Experimental Setup}
\label{subsec:exp_setup}

\noindent \textbf{Benchmarks and rewards.}
Following the experimental protocol of RAM and Flow-GRPO~\citep{liu2025flowgrpo}, we consider three representative text-to-image reward objectives: compositional correctness (GenEval), visual text rendering (OCR), and human preference alignment (PickScore). GenEval~\citep{ghosh2023geneval} measures whether generated images satisfy object, attribute, counting, and spatial-relation constraints. OCR evaluates visual text rendering through an edit-distance reward that checks whether the target text specified in the prompt appears legibly in the image. PickScore~\citep{kirstain2023pickapic} is a learned human-preference model trained from large-scale pairwise image comparisons.

For PickScore alignment, we use PickScore as the training reward. For OCR and GenEval, however, we use a multi-reward setting: OCR+PickScore and GenEval+PickScore, respectively. This design is motivated by \textbf{an empirical failure mode of prior single-reward optimization} (Fig.~\ref{fig:main_ocr}): although OCR or GenEval scores can become high, \textit{the generated images may collapse into reward-hacking artifacts, such as overly large text covering the whole image or simplified layouts that discard most visual details}. This collapse is reflected by DrawBench \textbf{quality metrics becoming even lower than the base model SD3.5M}. Adding PickScore as an auxiliary reward makes the optimization more conservative and better preserves general image quality while still improving the target task reward.

\noindent \textbf{Training protocol and AMD settings.}
We use Stable Diffusion 3.5 Medium (SD3.5M)~\citep{esser2024scaling} as the backbone, with separate teacher and student LoRA~\citep{hu2022lora} adapters ($r=32$, $\alpha=64$) for each reward setting and an additional fake-score adapter for four-step DMD. Training uses bf16 precision, 48 prompts per group, and 24 samples per prompt. The teacher learning rate is $3\times10^{-4}$. The student learning rate is $3\times10^{-4}$ for eight-step training and $1\times10^{-4}$ for four-step training; the fake-score learning rate is $3\times10^{-5}$. PickScore training uses 500/1000 iterations for 8/4-step REST-RAM and 1500 for REST-NFT; GenEval and OCR training uses 300 iterations. REST-RAM (REST by default) and REST-NFT denote students whose teacher branches are trained and evaluated following the official RAM and DiffusionNFT protocols, respectively. The student is jointly trained with AMD and evaluated using the EMA student (decay $0.9$) with 4/8 CFG-free steps. Auxiliary DMD loss is used in 4-step distillation. For AMD, each reward source uses group-relative advantages normalized within same-prompt samples to $[-1,1]$; OCR/GenEval task advantages are fused with PickScore advantages via Eq.~\ref{eq:multi_reward_adv}. We set the AMD scale to $\lambda=1$, shift to $b=0.5$, and KL-EMA coefficient to $0.2$. Training rewards are evaluated on held-out benchmark prompts, while generic image quality is evaluated on DrawBench~\citep{saharia2022photorealistic}.

\noindent \textbf{Generic image-quality evaluation.}
Optimizing a certain reward can degrade generic image quality, a failure mode often referred to as reward hacking. Following DiffusionNFT~\citep{diffusionnft}, we evaluate each trained model on DrawBench~\citep{saharia2022photorealistic} prompts that are disjoint from the reward-training and reward-test prompts. We report five quality and preference metrics: Aesthetic~\citep{schuhmann2022laion} and DeQA~\citep{you2025deqa} for perceptual quality, and ImageReward~\citep{xu2023imagereward}, HPSv2~\citep{wu2023human}, and PickScore~\citep{kirstain2023pickapic} for human preference. For the comparison in Tab.~\ref{tab:rtdmd}, we additionally report CLIPScore using CLIP ViT-L/14~\citep{radford2021clip}, computed as the mean paired image--text logit divided by $30$.

\begin{figure}[t]
  \centering
  \includegraphics[width=1\linewidth]{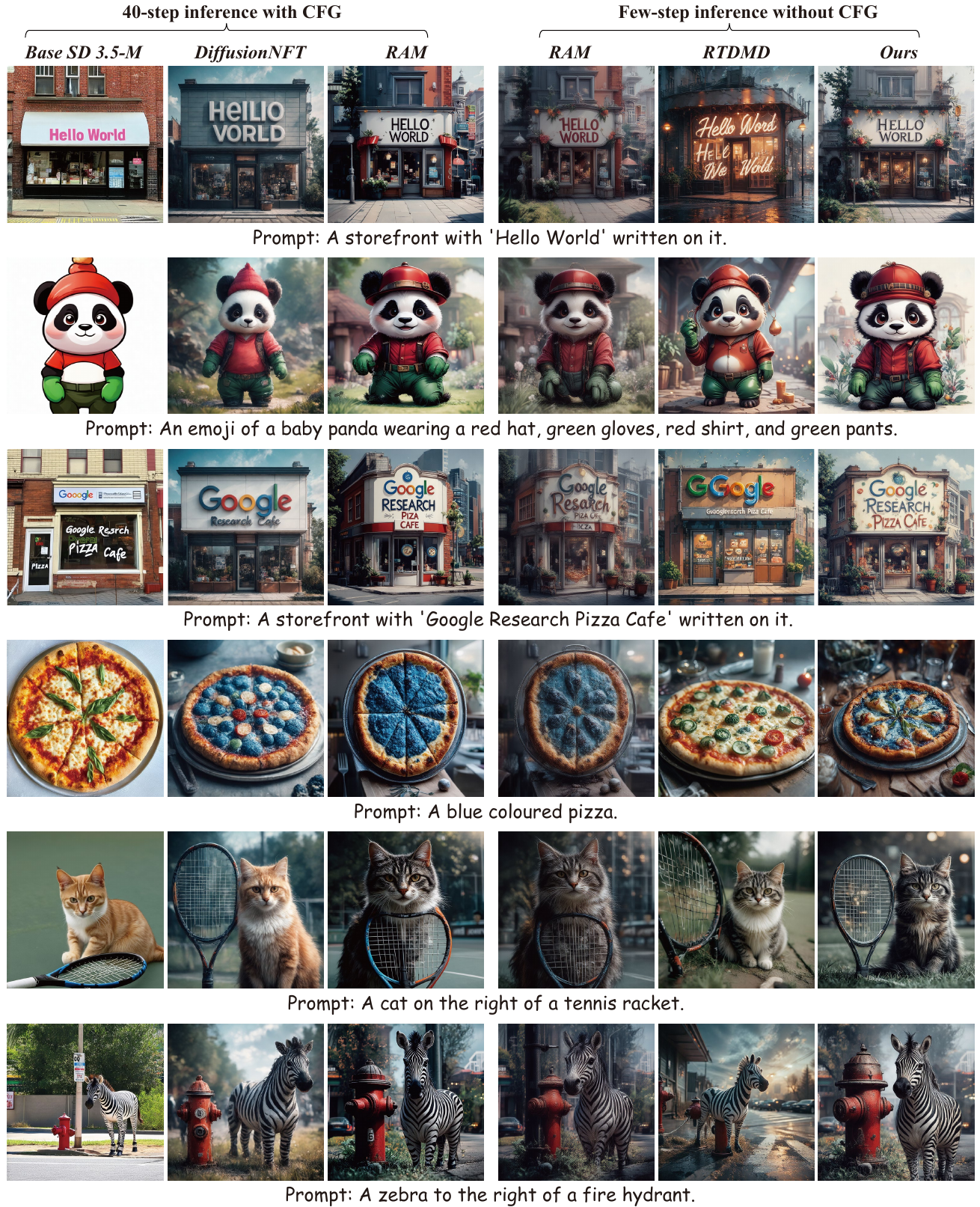}
  \caption{Qualitative comparison on DrawBench. REST achieves high-quality outputs with superior preference alignment under few-step inference, while matching full-step inference RL teacher.}
  \label{fig:main_drawbench}
\end{figure}

\subsection{Main Results}
\label{subsec:main_results}

\begin{table}[t]
  \centering
  
  \caption{SD3.5M post-training on GenEval, OCR, and PickScore. REST-RAM/NFT denote RAM/DiffusionNFT teachers. Gray rows use a single task reward; $\dagger$ adds PickScore.}

  \label{tab:sd35_results}
  \small
  \setlength{\tabcolsep}{4pt}
  \renewcommand{\arraystretch}{0.95}
  \begin{tabular}{@{}lccccccccc@{}}
    \toprule
    \textbf{Model} & \textbf{Steps} & \multicolumn{3}{c}{\textbf{Training Reward}} & \multicolumn{5}{c}{\textbf{Image Quality Metrics}} \\
    \cmidrule(lr){3-5}
    \cmidrule(lr){6-10}
    & & \textbf{GenEval} & \textbf{OCR} & \textbf{PickScore} & \textbf{Aesthetic} & \textbf{DeQA} & \textbf{ImgRwd} & \textbf{HPSv2} & \textbf{PickScore} \\
    \midrule
    SD3.5M & 40 & 0.64 & 0.63 & 21.78 & 5.39 & 4.08 & 0.85 & 0.28 & 22.40 \\
    SD3.5M & 8 & 0.28 & 0.12 & 19.50 & 5.15 & 2.38 & -0.96 & 0.18 & 20.39 \\
    \midrule
    \multicolumn{10}{c}{\textit{Compositional Image Generation}} \\
    \midrule
    Flow-GRPO & 40 & \cellcolor{gray!15}0.95 & \cellcolor{gray!15} & \cellcolor{gray!15} & \cellcolor{gray!15}5.25 & \cellcolor{gray!15}4.01 & \cellcolor{gray!15}1.03 & \cellcolor{gray!15}0.27 & \cellcolor{gray!15}22.37 \\
    AWM & 40 & \cellcolor{gray!15}0.83 & \cellcolor{gray!15} & \cellcolor{gray!15} & \cellcolor{gray!15}5.14 & \cellcolor{gray!15}3.75 & \cellcolor{gray!15}0.67 & \cellcolor{gray!15}0.24 & \cellcolor{gray!15}22.04 \\
    DiffusionNFT & 40 & \cellcolor{gray!15}0.95 & \cellcolor{gray!15} & \cellcolor{gray!15} & \cellcolor{gray!15}4.98 & \cellcolor{gray!15}4.10 & \cellcolor{gray!15}0.30 & \cellcolor{gray!15}0.24 & \cellcolor{gray!15}21.59 \\
    RAM & 40 & \cellcolor{gray!15}0.97 & \cellcolor{gray!15} & \cellcolor{gray!15} & \cellcolor{gray!15}5.38 & \cellcolor{gray!15}4.09 & \cellcolor{gray!15}1.19 & \cellcolor{gray!15}0.29 & \cellcolor{gray!15}22.52 \\
    RAM$^\dagger$ & 40 & \textbf{0.95} & & & 5.42 & 4.12 & \textbf{1.12} & 0.28 & 22.76 \\
    RAM$^\dagger$ & 8 & 0.65 & & & 5.46 & 2.27 & -0.19 & 0.20 & 21.36 \\
    \textbf{REST-RAM}$^\dagger$ & 8 & 0.94 & & & \textbf{5.67} & \textbf{4.15} & 1.05 & \textbf{0.30} & \textbf{22.79} \\
    \midrule
    \multicolumn{10}{c}{\textit{Visual Text Rendering}} \\
    \midrule
    Flow-GRPO & 40 & \cellcolor{gray!15} & \cellcolor{gray!15}0.92 & \cellcolor{gray!15} & \cellcolor{gray!15}5.32 & \cellcolor{gray!15}4.06 & \cellcolor{gray!15}0.95 & \cellcolor{gray!15}0.28 & \cellcolor{gray!15}22.44 \\
    AWM & 40 & \cellcolor{gray!15} & \cellcolor{gray!15}0.97 & \cellcolor{gray!15} & \cellcolor{gray!15}5.01 & \cellcolor{gray!15}2.83 & \cellcolor{gray!15}-0.85 & \cellcolor{gray!15}0.18 & \cellcolor{gray!15}20.56 \\
    DiffusionNFT & 40 & \cellcolor{gray!15} & \cellcolor{gray!15}0.96 & \cellcolor{gray!15} & \cellcolor{gray!15}4.87 & \cellcolor{gray!15}3.01 & \cellcolor{gray!15}-0.97 & \cellcolor{gray!15}0.18 & \cellcolor{gray!15}20.26 \\
    RAM & 40 & \cellcolor{gray!15} & \cellcolor{gray!15}0.97 & \cellcolor{gray!15} & \cellcolor{gray!15}5.23 & \cellcolor{gray!15}3.90 & \cellcolor{gray!15}0.44 & \cellcolor{gray!15}0.26 & \cellcolor{gray!15}21.83 \\
    RAM$^\dagger$ & 40 & & \textbf{0.96} & & 5.58 & \textbf{4.16} & \textbf{1.07} & \textbf{0.30} & 22.81 \\
    RAM$^\dagger$ & 8 & & 0.68 & & 5.59 & 2.60 & -0.21 & 0.22 & 21.45 \\
    \textbf{REST-RAM}$^\dagger$ & 8 & & \textbf{0.96} & & \textbf{5.65} & \textbf{4.16} & 1.00 & \textbf{0.30} & \textbf{22.83} \\
    \midrule
    \multicolumn{10}{c}{\textit{Human Preference Alignment}} \\
    \midrule
    Flow-GRPO & 40 & & & 23.31 & 5.92 & \textbf{4.22} & 1.28 & 0.32 & 23.53 \\
    AWM & 40 & & & 23.39 & 6.31 & 4.10 & 1.27 & 0.31 & 23.76 \\
    DiffusionNFT & 40 & & & 23.45 & 6.18 & 4.09 & 1.29 & 0.30 & 23.73 \\
    \textbf{REST-NFT} & 8 & & & 23.66 & 6.26 & 4.15 & 1.33 & 0.30 & 23.87 \\
    \textbf{REST-NFT} & 4 & & & 23.48 & 6.23 & 4.11 & 1.28 & 0.30 & 23.69 \\
    RAM & 40 & & & 23.70 & 6.17 & 4.21 & 1.35 & 0.32 & 23.95 \\
    RAM & 8 & & & 23.30 & \textbf{6.42} & 3.82 & 1.11 & 0.31 & 23.55 \\
    \textbf{REST-RAM} & 8 & & & \textbf{23.73} & 6.30 & 4.15 & 1.35 & \textbf{0.33} & 23.96 \\
    \textbf{REST-RAM} & 4 & & & 23.72 & 6.20 & 4.10 & \textbf{1.38} & 0.31 & \textbf{23.97} \\

    \bottomrule
  \end{tabular}
\end{table}

Tab.~\ref{tab:sd35_results} evaluates how well REST retains the gains of its corresponding 40-step RL teacher at 4/8 CFG-free steps. For GenEval and OCR, the gray-shaded rows report single-reward baselines as diagnostic references: these settings can achieve high task rewards while degrading generic image quality. We therefore focus on the rows marked by $\dagger$, which use the task reward together with PickScore and compare 8-step REST-RAM with its RAM teacher. The human-preference block additionally reports 4-step REST-RAM and both REST-NFT variants.

\begin{figure}[t]
  \centering
  \includegraphics[width=1\linewidth]{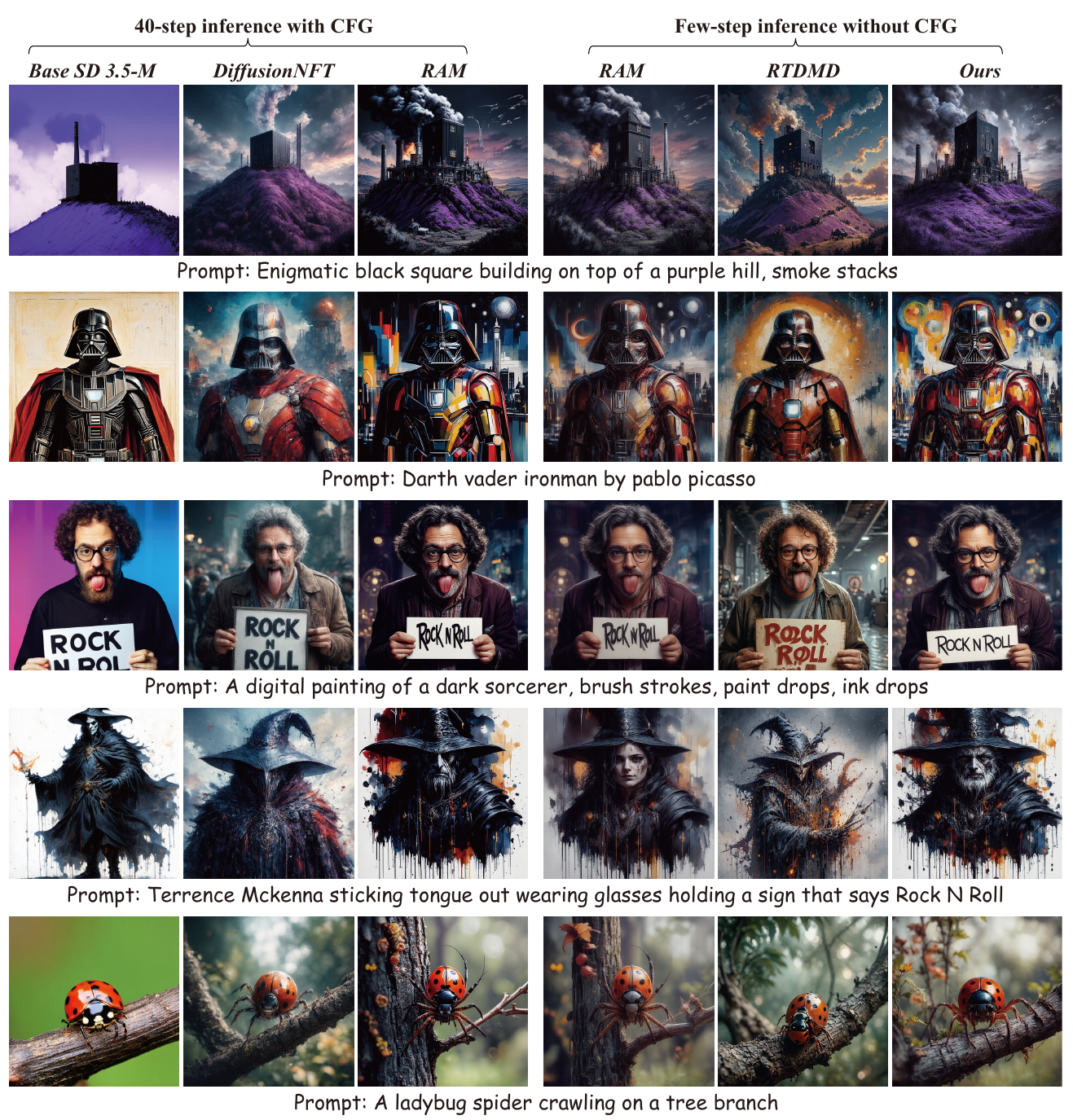}
  \caption{Qualitative comparison under PickScore alignment on the Pick-a-Pic testset.}
  \label{fig:main_pickscore}
\end{figure}

\begin{figure}[t]
  \centering
  \includegraphics[width=1\linewidth]{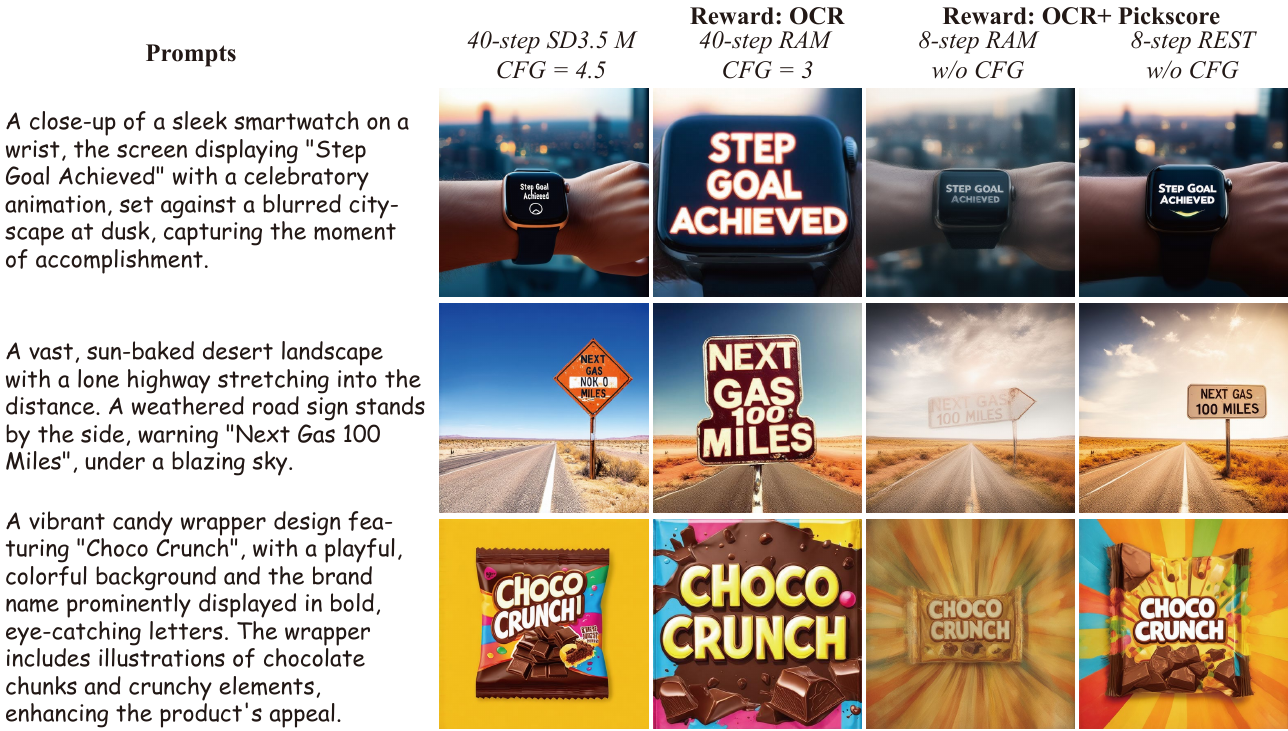}
  \vspace{-4mm}
  \caption{Qualitative comparison on visual text rendering. Single OCR reward leads to collapse by overly large text covering the whole image. OCR+PickScore optimization prevents the mode collapse, while REST achieves the highest image quality.}
  \label{fig:main_ocr}
\end{figure}

As shown in Tab.~\ref{tab:sd35_results}, REST achieves performance comparable to the 40-step RAM/DiffusionNFT teacher on both training rewards and out-of-domain DrawBench metrics, while using 4/8 CFG-free inference steps. It also obviously surpasses full-step RL baselines such as Flow-GRPO~\citep{liu2025flowgrpo}, AWM~\citep{awm} and DiffusionNFT~\citep{diffusionnft}. We note that 8-step RAM can occasionally obtain a high Aesthetic score, but this is often associated with fragmented structures and noisy details that hack the metric rather than reflect better perceptual quality. Overall, by continuously exposing the student to the evolving reward-optimized teacher rollouts and using AMD to select and amplify high-value trajectory segments, REST enables the student to go beyond plain trajectory compression. Qualitative comparisons in Figs.~\ref{fig:main_drawbench}, \ref{fig:main_pickscore}, and~\ref{fig:main_ocr} further show that REST preserves generic image quality, human preference, and visual text rendering under few-step CFG-free sampling.

\begin{table}[t]
  \centering
  \caption{4-step DrawBench comparison with RTDMD (1,500 warm-up + 1,000 parallel iterations). $\dagger$: official weights; $\ddagger$: reproduced with official code, PickScore reward, and Pick-a-Pic data.}
  \label{tab:rtdmd}
  \small
  \setlength{\tabcolsep}{5pt}
  \renewcommand{\arraystretch}{0.95}
  \begin{tabular}{@{}lccccccc@{}}
    \toprule
    \textbf{Methods} & \textbf{Steps} & \textbf{Train Iters} & \textbf{Aesthetic} & \textbf{ImgRwd} & \textbf{HPSv2} & \textbf{PickScore} & \textbf{CLIPScore} \\
    \midrule
    RTDMD$^\dagger$ & 4 & 1500+1000 & 6.09 & 1.26 & \textbf{0.326} & 23.30 & 0.930 \\
    RTDMD$^\ddagger$ & 4 & 1500+1000 & 6.18 & 1.27 & 0.310 & 23.71 & 0.885 \\
    \midrule
    \textbf{REST-RAM} & 4 & 1000 & \textbf{6.20} & \textbf{1.38} & 0.311 & \textbf{23.97} & \textbf{0.932} \\
    \bottomrule
  \end{tabular}
\end{table}

\noindent \textbf{Compared with Few-step RL Methods.}
DMDR and RTDMD~\citep{dmdr,rtdmd} combine a DMD warm-up stage with subsequent RL--DMD optimization. However, they use larger training datasets~\citep{schuhmann2021laion} and longer training of roughly 2,500--3,000 steps. More importantly, neither work directly compares its distilled generator against the corresponding full-step RL model trained under the same setting, leaving unclear how much of the RL policy's performance is actually preserved after distillation. REST instead targets the quality of full-step RL models under the popular GenEval, OCR, PickScore, and DrawBench protocols in diffusion RL fields, while requiring far fewer training steps. For an additional comparison, we evaluate the official RTDMD weights and reproduced RTDMD using official code on the same rewards and prompts as REST, at the training-independent DrawBench benchmark. As shown in Tab.~\ref{tab:rtdmd}, REST achieves stronger aesthetic and preference scores over both RTDMD versions despite using substantially fewer training steps.

\begin{figure}[t]
  \centering
  \includegraphics[width=1\linewidth]{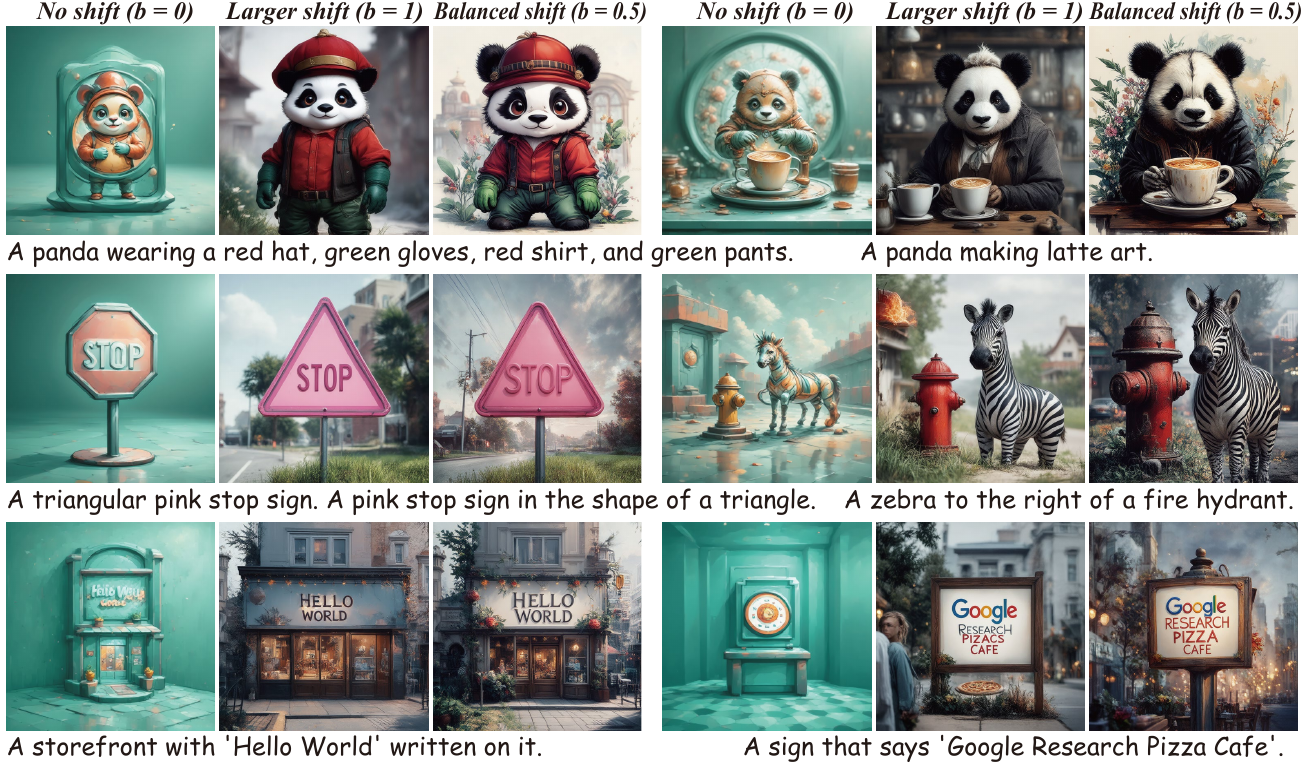}
  \vspace{-4mm}
  \caption{The shift $b$ controls the positive imitation prior in the AMD weight. A moderate shift preserves the necessary imitation prior while turning clearly low-quality trajectories into repulsive supervision, thereby amplifying the effect/style from RL signals.}
  \vspace{-4mm}
  \label{fig:ab_shift}
\end{figure}

\subsection{Ablation Study}
\label{subsec:ablation}

\begin{table}[t]
  \centering
  \caption{Progressive REST ablation: PickScore training, 8-step CFG-free evaluation on DrawBench.}
  \label{tab:ablation_progressive}
  \small
  \setlength{\tabcolsep}{4pt}
  \renewcommand{\arraystretch}{0.95}
  \begin{tabular}{@{}lcccccccc@{}}
    \toprule
    \textbf{Variant} & \textbf{Teacher RL} & \textbf{Dual Branch} & \textbf{AMD} & \textbf{Aesthetic} & \textbf{DeQA} & \textbf{ImgRwd} & \textbf{HPSv2} & \textbf{PickScore} \\
    \midrule
    Base model  & -- & -- & -- & 5.15 & 2.38 & -0.96 & 0.18 & 20.39 \\
    RL & Yes & -- & -- & \textbf{6.42} & 3.82 & 1.11 & 0.31 & 23.55 \\
    Distill then RL & -- & -- & -- & 5.83 & 3.54 & 1.19 & 0.28 & 23.14 \\
    RL then Distill & Yes & -- & -- & 6.22 & 4.14 & 1.32 & 0.32 & 23.83 \\
    RL \& Distill & Yes & Yes & -- & 6.20 & 4.09 & 1.32 & 0.32 & 23.74 \\
     \midrule
    \textbf{REST (full)} & Yes & Yes & Yes & 6.30 & \textbf{4.15} & \textbf{1.35} & \textbf{0.33} & \textbf{23.96} \\
    \bottomrule
  \end{tabular}
\end{table}

\noindent \textbf{Decoupled co-training and AMD.}
Tab.~\ref{tab:ablation_progressive} studies each component in 8-step REST-RAM PickScore training. We start from direct 8-step CFG-free inference with the base model, then add standard RL (RAM), a sequential RL-then-distill or distill-then-RL two-stage pipeline, simultaneous dual-branch RL with distillation, and finally the full REST model with AMD. The results lead to three observations. First, full-step standard RL itself improves 8-step CFG-free sampling, showing that reward optimization can partially compensate for the low-step generation gap. Second, simultaneous dual-branch RL-distillation in REST retains most of the gains of the RL-then-distill pipeline with far fewer training iterations, since the student reuses the teacher rollouts instead of requiring a separate distillation stage. Third, AMD further strengthens reward preference and improves perceptual quality, indicating that reward-aware trajectory weighting is more effective than uniform imitation.

\begin{table}[t]
  \centering
  \caption{Ablation of the AMD shift $b$ and EMA regularization under PickScore training, evaluated with DrawBench metrics.}
  \label{tab:ablation_shift_kl}
  \small
  \setlength{\tabcolsep}{5pt}
  \renewcommand{\arraystretch}{0.95}
  \begin{tabular}{@{}lccccccc@{}}
    \toprule
    \textbf{Variant} & \textbf{Shift $b$} & \textbf{EMA} & \textbf{Aesthetic} & \textbf{DeQA} & \textbf{ImgRwd} & \textbf{HPSv2} & \textbf{PickScore} \\
    \midrule
    No shift & 0 & Yes & 6.03 & 3.92 & 0.09 & 0.25 & 21.48 \\
    Larger shift & 1.0 & Yes & 6.25 & \textbf{4.19} & 1.33 & \underline{0.32} & 23.76 \\
    W/o EMA & 0.5 & No & \textbf{6.32} & 4.11 & \underline{1.34} & 0.31 & \underline{23.91} \\
    \midrule
    \textbf{REST (full)} & 0.5 & Yes & \underline{6.30} & \underline{4.15} & \textbf{1.35} & \textbf{0.33} & \textbf{23.96} \\
    \bottomrule
  \end{tabular}
\end{table}

\noindent \textbf{Shift and EMA regularization.}
Tab.~\ref{tab:ablation_shift_kl} further isolates the stabilizing components of the full AMD objective in the same PickScore setting. The shift $b$ controls the positive imitation prior in the AMD weight. Setting $b=0$ removes this baseline imitation pressure, making the objective dominated by aggressive negative modulation and thus potentially destabilizing early student training; a larger shift, in contrast, makes the objective closer to conservative imitation. As shown in Fig.~\ref{fig:ab_shift}, a moderate shift preserves the necessary imitation prior while turning clearly low-quality trajectories into repulsive supervision, thereby amplifying the effect of RL signals. We interpret this moderate negative modulation as a CFG-like contrastive force that is beneficial for improving generation quality. We also remove the EMA term to evaluate its stabilizing role: without this anchor, the student exhibits much stronger performance oscillation rather than a stable improvement, confirming that EMA regularization helps absorb the jitter caused by the continuously evolving teacher policy.

\begin{table}[t]
  \centering
  \caption{AMD/auxiliary DMD ablation: 4-step REST-RAM, PickScore training, DrawBench test.}
  \label{tab:ablation_dmd}
  \small
  \setlength{\tabcolsep}{5pt}
  \renewcommand{\arraystretch}{0.95}
  \begin{tabular}{@{}lccccc@{}}
    \toprule
    \textbf{Variant} & \textbf{Aesthetic} & \textbf{DeQA} & \textbf{ImgRwd} & \textbf{HPSv2} & \textbf{PickScore} \\
    \midrule
    DMD only & 6.10 & 3.99 & 1.38 & 0.30 & 23.58 \\
    AMD only & 6.16 & 4.04 & 1.33 & 0.31 & 23.93 \\
    On Student traj. & 6.26 & 4.08 & 1.35 & 0.33 & 23.95 \\
    \midrule
    \textbf{REST (full)} & 6.20 & 4.10 & 1.38 & 0.31 & 23.97 \\
    \bottomrule
  \end{tabular}
\end{table}

\noindent \textbf{AMD and auxiliary DMD for four-step generation.} Tab.~\ref{tab:ablation_dmd} isolates the contributions of AMD and auxiliary DMD in 4-step REST-RAM under PickScore training. AMD strengthens reward alignment, yielding higher PickScore, while DMD improves visual sharpness. We also compare DMD using student-rollout versus teacher-rollout states. Both yield similar results and are viable choices, while reusing teacher states avoids additional student rollouts and reduces sampling overhead.

\begin{figure}[t]
  \centering
  \includegraphics[width=1\linewidth]{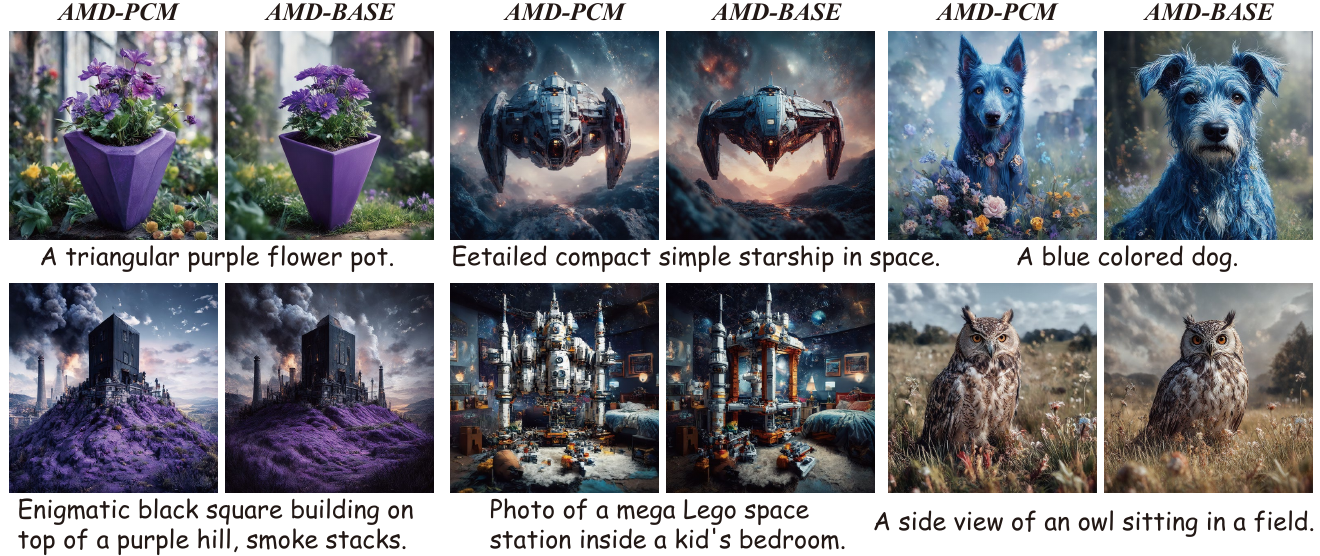}
  \caption{REST with different distillation objectives. AMD performs consistently with segment-velocity and PCM-style losses.}
  \label{fig:pcm}
\end{figure}

\subsection{Analysis}
\label{subsec:exp_analysis}

\noindent \textbf{Training efficiency.}
Since REST reuses the teacher's existing rollout trajectories and reward evaluations, it requires no additional rollout sampling, which is usually the most expensive stage in RL. Four-step REST-NFT and REST-RAM incur only \textbf{32.8\%} and \textbf{26.1\%} training-time overhead relative to standard DiffusionNFT and RAM training without distillation, respectively. Compared with sequential RL-then-distill training, REST achieves better performance with substantially fewer training iterations, as shown in Fig.~\ref{fig:motivation}.

\noindent \textbf{Generality across distillation losses.}
AMD only modulates the weight of a base distillation objective and is therefore not tied to the per-step imitation loss (segment-velocity MSE). We further integrate AMD with Phased-Consistency-Model-style phase consistency. Specifically, within each student-defined phase, REST trains every adjacent pair of teacher states to predict the same phase endpoint (more expensive than per-step imitation loss): the online student predicts from the higher-noise state, while the EMA student provides a stop-gradient target from the lower-noise state, and the two predictions are matched with a pseudo-Huber loss. The reward-derived AMD weight is then applied to this phase-local consistency loss. As shown in Fig.~\ref{fig:pcm}, the PCM variant surpasses 8-step RAM in our experiments, and gets similar performance with segment-velocity MSE, implying REST's compatibility with other distillation losses. Detailed procedures of the per-step imitation loss and PCM+AMD variants are provided in Algs.~\ref{alg:rest_base} and~\ref{alg:rest_pcm}.

\section{Conclusion}
\label{sec:conclusion}

We present \textbf{REST}, a unified RL-distillation co-training framework that turns reward-scored teacher trajectories into supervision for a few-step, CFG-free student. Its decoupled design preserves teacher RL optimization, while \textbf{Advantage-Modulated Distillation} (AMD) strengthens imitation of preferred trajectories and suppresses or repels low-reward ones. An auxiliary DMD loss using stored teacher states supports four-step generation. Experiments on SD3.5M demonstrate competitive reward alignment and image quality at four and eight inference steps. The ablation and analysis studies further validate the importance of decoupled co-training, advantage modulation, KL-EMA stabilization, and the generality of REST across teacher RL algorithms and distillation losses.

\bibliography{iclr2026_conference}

@String(CVPR  = {CVPR})

@String(ICLR  = {ICLR})

@inproceedings{radford2021clip,
  title={Learning transferable visual models from natural language supervision},
  author={Radford, Alec and Kim, Jong Wook and Hallacy, Chris and Ramesh, Aditya and Goh, Gabriel and Agarwal, Sandhini and Sastry, Girish and Askell, Amanda and Mishkin, Pamela and Clark, Jack and others},
  booktitle={Proc. ICML},
  year={2021}
}

@article{schuhmann2021laion,
  title={LAION-400M: Open Dataset of CLIP-Filtered 400 Million Image-Text Pairs},
  author={Christoph Schuhmann and Richard Vencu and Romain Beaumont and Robert Kaczmarczyk and Clayton Mullis and Aarush Katta and Theo Coombes and Jenia Jitsev and Aran Komatsuzaki},
  journal={arXiv:2111.02114},
  year={2021}
}

@article{song2020ddim,
  title={Denoising diffusion implicit models},
  author={Song, Jiaming and Meng, Chenlin and Ermon, Stefano},
  journal={arXiv preprint arXiv:2010.02502},
  year={2020}
}

@misc{flux,
    title = {FLUX.1-dev},
    howpublished = {\url{https://github.com/black-forest-labs/flux}},
    author = {Black-forest-labs},
    year = {2024}
}

@article{liu2025flowgrpo,
  title={Flow-grpo: Training flow matching models via online rl},
  author={Liu, Jie and Liu, Gongye and Liang, Jiajun and Li, Yangguang and Liu, Jiaheng and Wang, Xintao and Wan, Pengfei and Zhang, Di and Ouyang, Wanli},
  journal={arXiv preprint arXiv:2505.05470},
  year={2025}
}

@article{xue2025dancegrpo,
  title={Dancegrpo: Unleashing grpo on visual generation},
  author={Xue, Zeyue and Wu, Jie and Gao, Yu and Kong, Fangyuan and Zhu, Lingting and Chen, Mengzhao and Liu, Zhiheng and Liu, Wei and Guo, Qiushan and Huang, Weilin and others},
  journal={arXiv preprint arXiv:2505.07818},
  year={2025}
}

@misc{wu2025qwenimagetechnicalreport,
      title={Qwen-Image Technical Report}, 
      author={Chenfei Wu and Jiahao Li and Jingren Zhou and Junyang Lin and Kaiyuan Gao and Kun Yan and Sheng-ming Yin and Shuai Bai and Xiao Xu and Yilei Chen and Yuxiang Chen and Zecheng Tang and Zekai Zhang and Zhengyi Wang and An Yang and Bowen Yu and Chen Cheng and Dayiheng Liu and Deqing Li and Hang Zhang and Hao Meng and Hu Wei and Jingyuan Ni and Kai Chen and Kuan Cao and Liang Peng and Lin Qu and Minggang Wu and Peng Wang and Shuting Yu and Tingkun Wen and Wensen Feng and Xiaoxiao Xu and Yi Wang and Yichang Zhang and Yongqiang Zhu and Yujia Wu and Yuxuan Cai and Zenan Liu},
      year={2025}
}

@article{shao2024grpo,
  title={Deepseekmath: Pushing the limits of mathematical reasoning in open language models},
  author={Shao, Zhihong and Wang, Peiyi and Zhu, Qihao and Xu, Runxin and Song, Junxiao and Bi, Xiao and Zhang, Haowei and Zhang, Mingchuan and Li, YK and Wu, Yang and others},
  journal={arXiv preprint arXiv:2402.03300},
  year={2024}
}

@article{li2025mixgrpo,
  title={Mixgrpo: Unlocking flow-based grpo efficiency with mixed ode-sde},
  author={Li, Junzhe and Cui, Yutao and Huang, Tao and Ma, Yinping and Fan, Chun and Cheng, Yiming and Yang, Miles and Zhong, Zhao and Bo, Liefeng},
  journal={arXiv preprint arXiv:2507.21802},
  year={2025}
}

@article{xu2023imagereward,
  title={Imagereward: Learning and evaluating human preferences for text-to-image generation},
  author={Xu, Jiazheng and Liu, Xiao and Wu, Yuchen and Tong, Yuxuan and Li, Qinkai and Ding, Ming and Tang, Jie and Dong, Yuxiao},
  journal={NeurIPS},
  year={2023}
}

@article{hu2022lora,
  title={Lora: Low-rank adaptation of large language models.},
  author={Hu, Edward J and Shen, Yelong and Wallis, Phillip and Allen-Zhu, Zeyuan and Li, Yuanzhi and Wang, Shean and Wang, Lu and Chen, Weizhu and others},
  journal={ICLR},
  year={2022}
}

@article{lipman2022flow,
  title={Flow matching for generative modeling},
  author={Lipman, Yaron and Chen, Ricky TQ and Ben-Hamu, Heli and Nickel, Maximilian and Le, Matt},
  journal={arXiv preprint arXiv:2210.02747},
  year={2022}
}

@inproceedings{leapalign,
  title={LeapAlign: Post-Training Flow Matching Models at Any Generation Step by Building Two-Step Trajectories},
  author={Liang, Zhanhao and Yang, Tao and Wu, Jie and Feng, Chengjian and Zheng, Liang},
  booktitle={Proceedings of the IEEE/CVF Conference on Computer Vision and Pattern Recognition},
  pages={23238--23248},
  year={2026}
}

@article{rtdmd,
  title={Reinforcing Few-step Generators via Reward-Tilted Distribution Matching},
  author={Huang, Yushi and Zhou, Xiangxin and Wang, Ruoyu and Zhang, Chi and Zhang, Jun and Pang, Tianyu},
  journal={arXiv preprint arXiv:2605.26108},
  year={2026}
}

@inproceedings{dmd,
  title={One-step diffusion with distribution matching distillation},
  author={Yin, Tianwei and Gharbi, Micha{\"e}l and Zhang, Richard and Shechtman, Eli and Durand, Fredo and Freeman, William T and Park, Taesung},
  booktitle={Proceedings of the IEEE/CVF conference on computer vision and pattern recognition},
  pages={6613--6623},
  year={2024}
}

@article{dmd2,
  title={Improved distribution matching distillation for fast image synthesis},
  author={Yin, Tianwei and Gharbi, Micha{\"e}l and Park, Taesung and Zhang, Richard and Shechtman, Eli and Durand, Fredo and Freeman, William T},
  journal={Advances in neural information processing systems},
  volume={37},
  pages={47455--47487},
  year={2024}
}

@article{dmdr,
  title={Distribution Matching Distillation Meets Reinforcement Learning},
  author={Jiang, Dengyang and Liu, Dongyang and Wang, Zanyi and Wu, Qilong and Li, Liuzhuozheng and Li, Hengzhuang and Jin, Xin and Liu, David and Li, Zhen and Zhang, Bo and others},
  journal={arXiv preprint arXiv:2511.13649},
  year={2025}
}

@article{geneval,
  title={Geneval: An object-focused framework for evaluating text-to-image alignment},
  author={Ghosh, Dhruba and Hajishirzi, Hannaneh and Schmidt, Ludwig},
  journal={Advances in Neural Information Processing Systems},
  volume={36},
  pages={52132--52152},
  year={2023}
}

@article{dpo,
  title={Direct preference optimization: Your language model is secretly a reward model},
  author={Rafailov, Rafael and Sharma, Archit and Mitchell, Eric and Manning, Christopher D and Ermon, Stefano and Finn, Chelsea},
  journal={Advances in neural information processing systems},
  volume={36},
  pages={53728--53741},
  year={2023}
}

@inproceedings{diffusiondpo,
  title={Diffusion model alignment using direct preference optimization},
  author={Wallace, Bram and Dang, Meihua and Rafailov, Rafael and Zhou, Linqi and Lou, Aaron and Purushwalkam, Senthil and Ermon, Stefano and Xiong, Caiming and Joty, Shafiq and Naik, Nikhil},
  booktitle={Proceedings of the IEEE/CVF Conference on Computer Vision and Pattern Recognition},
  pages={8228--8238},
  year={2024}
}

@inproceedings{spo,
  title={Aesthetic post-training diffusion models from generic preferences with step-by-step preference optimization},
  author={Liang, Zhanhao and Yuan, Yuhui and Gu, Shuyang and Chen, Bohan and Hang, Tiankai and Cheng, Mingxi and Li, Ji and Zheng, Liang},
  booktitle={Proceedings of the IEEE/CVF Conference on Computer Vision and Pattern Recognition},
  pages={13199--13208},
  year={2025}
}

@article{diffusionnft,
  title={Diffusionnft: Online diffusion reinforcement with forward process},
  author={Zheng, Kaiwen and Chen, Huayu and Ye, Haotian and Wang, Haoxiang and Zhang, Qinsheng and Jiang, Kai and Su, Hang and Ermon, Stefano and Zhu, Jun and Liu, Ming-Yu},
  journal={arXiv preprint arXiv:2509.16117},
  year={2025}
}

@article{awm,
  title={Advantage weighted matching: Aligning rl with pretraining in diffusion models},
  author={Xue, Shuchen and Ge, Chongjian and Zhang, Shilong and Li, Yichen and Ma, Zhi-Ming},
  journal={arXiv preprint arXiv:2509.25050},
  year={2025}
}

@article{pcm,
  title={Phased consistency models},
  author={Wang, Fu-Yun and Huang, Zhaoyang and Bergman, Alexander W and Shen, Dazhong and Gao, Peng and Lingelbach, Michael and Sun, Keqiang and Bian, Weikang and Song, Guanglu and Liu, Yu and others},
  journal={Advances in neural information processing systems},
  volume={37},
  pages={83951--84009},
  year={2024}
}

@inproceedings{cm,
  title={Consistency models},
  author={Song, Yang and Dhariwal, Prafulla and Chen, Mark and Sutskever, Ilya},
  booktitle={Proceedings of the 40th International Conference on Machine Learning},
  pages={32211--32252},
  year={2023}
}

@article{lcm,
  title={Latent consistency models: Synthesizing high-resolution images with few-step inference},
  author={Luo, Simian and Tan, Yiqin and Huang, Longbo and Li, Jian and Zhao, Hang},
  journal={arXiv preprint arXiv:2310.04378},
  year={2023}
}

@article{sim,
  title={One-step diffusion distillation through score implicit matching},
  author={Luo, Weijian and Huang, Zemin and Geng, Zhengyang and Kolter, J Zico and Qi, Guo-jun},
  journal={Advances in Neural Information Processing Systems},
  volume={37},
  pages={115377--115408},
  year={2024}
}

@inproceedings{drtune,
  title={Deep reward supervisions for tuning text-to-image diffusion models},
  author={Wu, Xiaoshi and Hao, Yiming and Zhang, Manyuan and Sun, Keqiang and Huang, Zhaoyang and Song, Guanglu and Liu, Yu and Li, Hongsheng},
  booktitle={European Conference on Computer Vision},
  pages={108--124},
  year={2024},
  organization={Springer}
}

@inproceedings{draft,
  title={Directly fine-tuning diffusion models on differentiable rewards},
  author={Clark, Kevin and Vicol, Paul and Swersky, Kevin and Fleet, David},
  booktitle={International Conference on Learning Representations},
  volume={2024},
  pages={4793--4822},
  year={2024}
}

@article{progressive_distil,
  title={Progressive distillation for fast sampling of diffusion models},
  author={Salimans, Tim and Ho, Jonathan},
  journal={arXiv preprint arXiv:2202.00512},
  year={2022}
}

@inproceedings{swiftbrush,
  title={Swiftbrush: One-step text-to-image diffusion model with variational score distillation},
  author={Nguyen, Thuan Hoang and Tran, Anh},
  booktitle={Proceedings of the IEEE/CVF Conference on Computer Vision and Pattern Recognition},
  pages={7807--7816},
  year={2024}
}

@article{bergmeister2026ram,
  title={Reinforce Adjoint Matching: Scaling RL Post-Training of Diffusion and Flow-Matching Models},
  author={Bergmeister, Andreas and Jegelka, Stefanie and N{\"u}sken, Nikolas and Domingo-Enrich, Carles and Pidstrigach, Jakiw},
  journal={arXiv preprint arXiv:2605.10759},
  year={2026}
}

@article{ho2022cfg,
  title={Classifier-free diffusion guidance},
  author={Ho, Jonathan and Salimans, Tim},
  journal={arXiv preprint arXiv:2207.12598},
  year={2022}
}

@article{fan2026rdm,
  title={Rdm: Re-conceptualizing Distribution Matching as a Reward for Diffusion Distillation},
  author={Fan, Linqian and Sun, Peiqin and Wen, Tiancheng and Lu, Shun and Song, Chengru},
  journal={arXiv preprint arXiv:2603.28460},
  year={2026}
}

@inproceedings{peters2007rwr,
  title={Reinforcement learning by reward-weighted regression for operational space control},
  author={Peters, Jan and Schaal, Stefan},
  booktitle={Proceedings of the 24th international conference on Machine learning},
  pages={745--750},
  year={2007}
}

@article{peng2019awr,
  title={Advantage-weighted regression: Simple and scalable off-policy reinforcement learning},
  author={Peng, Xue Bin and Kumar, Aviral and Zhang, Grace and Levine, Sergey},
  journal={arXiv preprint arXiv:1910.00177},
  year={2019}
}

@article{kostrikov2021iql,
  title={Offline reinforcement learning with implicit q-learning},
  author={Kostrikov, Ilya and Nair, Ashvin and Levine, Sergey},
  journal={arXiv preprint arXiv:2110.06169},
  year={2021}
}

@article{luhman2021knowledge,
  title={Knowledge distillation in iterative generative models for improved sampling speed},
  author={Luhman, Eric and Luhman, Troy},
  journal={arXiv preprint arXiv:2101.02388},
  year={2021}
}

@inproceedings{ghosh2023geneval,
  title={GenEval: An Object-Focused Framework for Evaluating Text-to-Image Alignment},
  author={Ghosh, Dhruba and Hajishirzi, Hanna and Schmidt, Ludwig},
  booktitle={Advances in Neural Information Processing Systems},
  volume={36},
  year={2023},
}

@inproceedings{kirstain2023pickapic,
  title={Pick-a-Pic: An Open Dataset of User Preferences for Text-to-Image Generation},
  author={Yuval Kirstain and Adam Polyak and Uriel Singer and Shahbuland Matiana and Joe Penna and Omer Levy},
  booktitle={Advances in Neural Information Processing Systems},
  volume={36},
  year={2023},
}

@article{wu2023human,
  title={Human Preference Score v2: A Solid Benchmark for Evaluating Human Preferences of Text-to-Image Synthesis},
  author={Wu, Xiaoshi and Hao, Yiming and Sun, Keqiang and Chen, Yixiong and Zhu, Feng and Zhao, Rui and Li, Hongsheng},
  journal={arXiv preprint arXiv:2306.09341},
  year={2023},
}

@misc{schuhmann2022laion,
  title={LAION-aesthetics},
  author={Christoph Schuhmann and Romain Beaumont},
  howpublished={laion.ai},
  year={2022},
}

@inproceedings{saharia2022photorealistic,
  title={Photorealistic Text-to-Image Diffusion Models with Deep Language Understanding},
  author={Saharia, Chitwan and Chan, William and Saxena, Saurabh and Li, Lala and Whang, Jay and Denton, Emily L. and Ghasemipour, Seyed Kamyar Seyed and Gontijo Lopes, Raphael and Karagol Ayan, Burcu and Salimans, Tim and Ho, Jonathan and Fleet, David J. and Norouzi, Mohammad},
  booktitle={Advances in Neural Information Processing Systems},
  volume={35},
  pages={36479--36494},
  year={2022},
}

@inproceedings{you2025deqa,
  title={Teaching Large Language Models to Regress Accurate Image Quality Scores Using Score Distribution},
  author={You, Zhiyuan and Cai, Xin and Gu, Jinjin and Xue, Tianfan and Dong, Chao},
  booktitle={Proceedings of the IEEE/CVF Conference on Computer Vision and Pattern Recognition (CVPR)},
  pages={14483--14494},
  year={2025},
}

@inproceedings{esser2024scaling,
  title={Scaling Rectified Flow Transformers for High-Resolution Image Synthesis},
  author={Esser, Patrick and Kulal, Sumith and Blattmann, Andreas and Entezari, Rahim and M{\"u}ller, Jonas and Saini, Harry and Levi, Yam and Lorenz, Dominik and Sauer, Axel and Boesel, Frederic and Podell, Dustin and Dockhorn, Tim and English, Zion and Rombach, Robin},
  booktitle={Proceedings of the 41st International Conference on Machine Learning},
  series={Proceedings of Machine Learning Research},
  volume={235},
  pages={12606--12633},
  publisher={PMLR},
  year={2024}
}

@inproceedings{chen2026nft,
  title={NFT: Bridging supervised learning and reinforcement learning in math reasoning},
  author={Chen, Huayu and Zheng, Kaiwen and Zhang, Qinsheng and Cui, Ganqu and Cui, Yin and Ye, Haotian and Lin, Tsung-Yi and Liu, Ming-Yu and Zhu, Jun and Wang, Haoxiang},
  booktitle={International Conference on Learning Representations},
  volume={2026},
  pages={124025--124042},
  year={2026}
}
\bibliographystyle{iclr2026_conference}

\clearpage
\appendix
\section{Relationship between DiffusionNFT and RAM}
\label{app:nft_ram_similarity}

Our experiments instantiate REST with both RAM and DiffusionNFT teachers. REST decouples teacher-side reinforcement learning from student-side distillation: the student consumes reward-scored ODE trajectories without depending on the teacher's specific policy-optimization objective. This section complements the empirical comparison by examining the forward-process objectives of RAM and DiffusionNFT. We show that these representative methods share an anchored, reward-shifted velocity-regression structure despite differences in their anchor choices and hyperparameters. Rather than claiming that the two algorithms are fully equivalent, this structural commonality identifies a shared trajectory interface that enables REST to accommodate different teacher optimizers without changing its student-side distillation mechanism.

To make this connection explicit, we compare their objectives on the forward noising process. Let $v^{\rm gt}=\epsilon-x_0$ denote the flow-matching target, $v^{\rm old}$ the lagged policy used for sampling, and $v_\phi$ the trainable policy. DiffusionNFT constructs two implicit policies symmetric around $v^{\rm old}$,
\begin{equation}
    v_\phi^+=(1-\beta)v^{\rm old}+\beta v_\phi,
    \qquad
    v_\phi^-=(1+\beta)v^{\rm old}-\beta v_\phi,
    \label{eq:app_nft_policies}
\end{equation}
and optimizes the reward-weighted flow-matching objective
\begin{equation}
    \mathcal{L}_{\rm NFT}
    =r\left\|v_\phi^+-v^{\rm gt}\right\|^2
    +(1-r)\left\|v_\phi^--v^{\rm gt}\right\|^2,
    \label{eq:app_nft_core}
\end{equation}
where $r\in[0,1]$ is obtained from the clipped group-relative advantage. For clarity, we first omit the time-dependent and detached self-normalization factors used in the implementation; their effect is discussed below.

\paragraph{Equivalent target of DiffusionNFT.}
Define
\begin{equation}
    \Delta=v^{\rm old}-v^{\rm gt},
    \qquad
    \delta_\phi=v_\phi-v^{\rm old},
    \qquad
    A=2r-1\in[-1,1].
\end{equation}
Equation~\ref{eq:app_nft_policies} gives
$v_\phi^+-v^{\rm gt}=\Delta+\beta\delta_\phi$ and
$v_\phi^--v^{\rm gt}=\Delta-\beta\delta_\phi$. Substituting them into Eq.~\ref{eq:app_nft_core} yields
\begin{align}
    \mathcal{L}_{\rm NFT}
    &=r\left\|\Delta+\beta\delta_\phi\right\|^2
      +(1-r)\left\|\Delta-\beta\delta_\phi\right\|^2 \\
    &=\left\|\Delta\right\|^2
      +2\beta A\langle\Delta,\delta_\phi\rangle
      +\beta^2\left\|\delta_\phi\right\|^2 \\
    &=\beta^2\left\|\delta_\phi+\frac{A}{\beta}\Delta\right\|^2
      +(1-A^2)\left\|\Delta\right\|^2.
    \label{eq:app_nft_square}
\end{align}
The last term is independent of $v_\phi$. Therefore, Eq.~\ref{eq:app_nft_core} is gradient-equivalent to
\begin{equation}
    \widetilde{\mathcal{L}}_{\rm NFT}
    =\beta^2\left\|v_\phi-v_{\rm target}^{\rm NFT}\right\|^2,
    \qquad
    v_{\rm target}^{\rm NFT}
    =v^{\rm old}+\frac{A}{\beta}
      \left(v^{\rm gt}-v^{\rm old}\right).
    \label{eq:app_nft_target}
\end{equation}
Thus, DiffusionNFT can be interpreted as regression from the old policy toward a reward-dependent target. Positive advantages move the policy toward $v^{\rm gt}$, while negative advantages extrapolate it away from $v^{\rm gt}$ through the same anchor $v^{\rm old}$.

\paragraph{Comparison with RAM.}
Following the RAM formulation in Eq.~\ref{eq:ram_loss}, the stop-gradient target is
\begin{equation}
    v_{\rm target}^{\rm RAM}
    =v^{\rm base}+\eta A_{\rm RAM}
      \left(v^{\rm gt}-v^{\rm old}\right),
    \qquad
    \mathcal{L}_{\rm RAM}
    =\left\|v_\phi-\operatorname{sg}
      \left(v_{\rm target}^{\rm RAM}\right)\right\|^2,
    \label{eq:app_ram_target}
\end{equation}
where $v^{\rm base}$ is the frozen pretrained velocity, $A_{\rm RAM}$ is the normalized advantage, and $\eta$ is its scale. Equations~\ref{eq:app_nft_target} and~\ref{eq:app_ram_target} expose their shared template,
\begin{equation}
    v_{\rm target}
    =v_{\rm anchor}+\gamma(A)
      \left(v^{\rm gt}-v_{\rm res}\right).
    \label{eq:app_unified_target}
\end{equation}
Here $v_{\rm res}$ denotes the velocity used in the reward-scaled residual. DiffusionNFT uses $v_{\rm anchor}=v_{\rm res}=v^{\rm old}$ and $\gamma(A)=A/\beta$, whereas RAM uses $v_{\rm anchor}=v^{\rm base}$, $v_{\rm res}=v^{\rm old}$, and $\gamma(A)=\eta A_{\rm RAM}$. Their targets approximately coincide when $v^{\rm old}\approx v^{\rm base}$ and $A/\beta\approx\eta A_{\rm RAM}$.

\paragraph{Full DiffusionNFT objective and limitations of the equivalence.}
The implementation of DiffusionNFT uses detached residual normalizers $w_+$ and $w_-$ and an additional time weight. With the definitions above, its gradient has the form
\begin{equation}
    \nabla_{v_\phi}\mathcal{L}_{\rm NFT}
    \propto t\left[
    \left(\frac{r}{w_+}-\frac{1-r}{w_-}\right)\Delta
    +\beta\left(\frac{r}{w_+}+\frac{1-r}{w_-}\right)\delta_\phi
    \right].
    \label{eq:app_nft_full_gradient}
\end{equation}
Because $w_+$ and $w_-$ are stop-gradient quantities, this remains an anchored regression update: the first term supplies the reward-dependent direction and the second pulls the trainable policy toward $v^{\rm old}$. When $w_+=w_-$, Eq.~\ref{eq:app_nft_full_gradient} reduces exactly, up to a positive scalar, to the gradient of Eq.~\ref{eq:app_nft_target}. When the two normalizers are merely close, this equivalence is approximate; when they differ substantially, they adaptively rescale the effective advantage and trust-region strength.

The relationship above establishes structural similarity rather than complete equivalence. DiffusionNFT parameterizes symmetric positive and negative policies, applies time-dependent self-normalization, and anchors its policy-improvement target at $v^{\rm old}$. RAM also uses the lagged policy $v^{\rm old}$ in the reward-scaled residual, but anchors its stop-gradient target at the frozen $v^{\rm base}$ and uses a separately chosen advantage scale. Both methods incorporate reward feedback into forward-process velocity regression. This shared structure complements the observed compatibility of REST with RAM and DiffusionNFT. The trajectory interface, rather than equivalence of their optimization objectives, enables reuse of the student-side distillation mechanism.

\section{Detailed Training Algorithms of REST}
\label{app:rest_algorithms}

We provide training procedures for the two base trajectory distillation objectives, segment-velocity regression and PCM-style consistency. Both reuse the teacher trajectories and terminal rewards. The additional DMD update used for four-step generation is described below. Let $\phi$, $\bar\phi$, $\theta$, and $\bar\theta$ denote the online teacher, EMA teacher, online student, and EMA student, respectively. Let $M$ and $K$ be the numbers of teacher and student steps, with $K<M$.

\begin{algorithm}[htbp]
\caption{REST with Segment-Velocity Distillation and AMD (Base Objective)}
\label{alg:rest_base}
\small
\begin{algorithmic}[1]
\Require Online/EMA teacher $v_\phi,v_{\bar\phi}$; online/EMA student $v_\theta,v_{\bar\theta}$; rewards $\{R_m\}_{m=1}^{N_R}$.
\Require Teacher steps $M$; student boundaries $0=q_0<\cdots<q_K=M$; AMD parameters $\lambda,b,\{\alpha_m\}$; EMA coefficient $\beta_{\rm ema}$.
\Statex \hspace{-\algorithmicindent}{\textbf{while} not converged}
    \Statex \textbf{Phase 1: Teacher Rollout and Multi-Reward Advantage}
    \State Sample prompt $c$ and a complete trajectory $\mathcal{T}=\{x_{t_0},\ldots,x_{t_M}\}$ with $v_{\bar\phi}$, using the teacher algorithm's solver and guidance settings.
    \State Decode $x_{t_M}$ and compute $R_m(x_{t_M},c)$ for every reward source $m$.
    \State Normalize and clip each reward within the same-prompt group to obtain $A^{(m)}\in[-1,1]$.
    \State Fuse advantages: $A_{\rm mix}=\sum_m\alpha_m A^{(m)}/\sum_m\alpha_m$.

    \Statex \textbf{Phase 2: Decoupled Teacher RL Update}
    \State Update $\phi$ using the original teacher RL objective and reward preprocessing on the terminal samples and rewards.
    \State Do not propagate student gradients into $\phi$ or the stored trajectory $\mathcal{T}$.

    \Statex \textbf{Phase 3: Segment-Velocity Student Distillation}
    \State Initialize $\mathcal{L}_{\rm student}^{\rm seg}\leftarrow 0$.
    \For{$k=0,\ldots,K-1$}
        \State Construct $s_k=(x_{t_{q_k}},t_{q_k},c)$ and segment target
        \Statex \hspace{\algorithmicindent}$a_k=(x_{t_{q_{k+1}}}-x_{t_{q_k}})/(t_{q_{k+1}}-t_{q_k})$.
        \State Compute $w_{\rm AMD}=\lambda(A_{\rm mix}+b)$.
        \State Compute $\ell_{\rm seg}=\|v_\theta(s_k)-\operatorname{sg}(a_k)\|^2$.
        \State Compute $\ell_{\rm ema}=\|v_\theta(s_k)-\operatorname{sg}(v_{\bar\theta}(s_k))\|^2$.
        \State Accumulate $\mathcal{L}_{\rm student}^{\rm seg}\mathrel{+}=w_{\rm AMD}\ell_{\rm seg}+\beta_{\rm ema}\ell_{\rm ema}$.
    \EndFor

    \Statex \textbf{Phase 4: Independent Optimization and EMA Update}
    \State Update only $\theta$ by minimizing $\mathcal{L}_{\rm student}^{\rm seg}$.
    \State $\bar\phi\leftarrow\rho_T\bar\phi+(1-\rho_T)\phi$; $\bar\theta\leftarrow\rho_S\bar\theta+(1-\rho_S)\theta$.
\Statex \hspace{-\algorithmicindent}{\textbf{end while}}
\Ensure EMA student $v_{\bar\theta}$ for $K$-step CFG-free inference.
\end{algorithmic}
\end{algorithm}

\noindent\textbf{Preliminary: PCM-style phase consistency.}
The default REST variant above represents each student step by a single segment velocity between two selected teacher states. The PCM variant instead divides the fine-grained $M$-step teacher trajectory into $K$ student phases and enforces consistency within each phase. The indices $q_k$ and $q_{k+1}$ denote the teacher-grid columns at the beginning and end of student phase $k$, respectively. Within this phase, $j\in\{q_k,\ldots,q_{k+1}-1\}$ indexes an adjacent pair of teacher states. Since the denoising schedule satisfies $t_j>t_{j+1}$, $x_{t_j}$ is the higher-noise state and $x_{t_{j+1}}$ is the immediately following lower-noise state. All adjacent pairs in phase $k$ share the same endpoint time $t_{q_{k+1}}$.

Starting from the higher-noise state, the online student predicts the phase-end latent as $\widehat{x}^{\rm on}_{k,j}=x_{t_j}+(t_{q_{k+1}}-t_j)v_\theta(x_{t_j},t_j,c)$. Here, the superscript ``on'' denotes the online, trainable student, and the hat indicates that this is a predicted endpoint latent rather than an observed teacher state. From the lower-noise state, the EMA student analogously produces $\widehat{x}^{\rm ema}_{k,j}$, where ``ema'' denotes the lagged student and the prediction is treated as a stop-gradient target. In these expressions, $v_\theta(\cdot,t,c)$ is the conditional velocity predicted at time $t$ for prompt $c$. PCM trains the two endpoint predictions to agree through the pseudo-Huber distance $\ell_{\rm pcm}^{k,j}$. Intuitively, regardless of which nearby state within a phase is used as the starting point, the student should predict the same phase endpoint. REST then applies the rollout-level AMD weight $w_{\rm AMD}=\lambda(A_{\rm mix}+b)$ to this consistency loss, strengthening phase-consistency learning on preferred trajectories and weakening or reversing it on low-reward trajectories.

\begin{algorithm}[htbp]
\caption{REST with PCM-Style Phase Consistency and AMD}
\label{alg:rest_pcm}
\small
\begin{algorithmic}[1]
\Require Online/EMA teacher $v_\phi,v_{\bar\phi}$; online/EMA student $v_\theta,v_{\bar\theta}$; rewards $\{R_m\}_{m=1}^{N_R}$.
\Require Teacher steps $M$; phase boundaries $0=q_0<\cdots<q_K=M$; AMD parameters $\lambda,b,\{\alpha_m\}$; EMA coefficient $\beta_{\rm ema}$; pseudo-Huber constant $c_{\rm huber}$.
\Statex \hspace{-\algorithmicindent}{\textbf{while} not converged}
    \Statex \textbf{Phase 1: Teacher Rollout and Multi-Reward Advantage}
    \State Sample prompt $c$ and a complete trajectory $\mathcal{T}=\{x_{t_0},\ldots,x_{t_M}\}$ with $v_{\bar\phi}$, using the teacher algorithm's solver and guidance settings.
    \State Decode $x_{t_M}$, compute all rewards, and obtain the fused advantage
    \Statex \hspace{\algorithmicindent}$A_{\rm mix}=\sum_m\alpha_m A^{(m)}/\sum_m\alpha_m$ after per-reward normalization and clipping.

    \Statex \textbf{Phase 2: Decoupled Teacher RL Update}
    \State Update $\phi$ with its original RL objective; detach $\mathcal{T}$ from the teacher graph.

    \Statex \textbf{Phase 3: PCM+AMD Student Distillation}
    \State Initialize $\mathcal{L}_{\rm student}^{\rm pcm}\leftarrow 0$.
    \For{$k=0,\ldots,K-1$}
        \For{$j=q_k,\ldots,q_{k+1}-1$}
            \State Set the common phase endpoint to $t_{q_{k+1}}$.
            \State Predict the endpoint from the higher-noise state $x_{t_j}$ with the online student:
            \Statex \hspace{\algorithmicindent}$\widehat{x}^{\rm on}_{k,j}=x_{t_j}+(t_{q_{k+1}}-t_j)v_\theta(x_{t_j},t_j,c)$.
            \State Predict the same endpoint from the lower-noise state $x_{t_{j+1}}$ with the EMA student:
            \Statex \hspace{\algorithmicindent}$\widehat{x}^{\rm ema}_{k,j}=\operatorname{sg}[x_{t_{j+1}}+(t_{q_{k+1}}-t_{j+1})v_{\bar\theta}(x_{t_{j+1}},t_{j+1},c)]$.
            \State Compute $\ell_{\rm pcm}^{k,j}=\sqrt{\|\widehat{x}^{\rm on}_{k,j}-\widehat{x}^{\rm ema}_{k,j}\|_2^2+c_{\rm huber}^2}-c_{\rm huber}$.
            \State Compute $w_{\rm AMD}=\lambda(A_{\rm mix}+b)$.
            \State Compute $\ell_{\rm ema}^{k,j}=\|v_\theta(x_{t_j},t_j,c)-\operatorname{sg}(v_{\bar\theta}(x_{t_j},t_j,c))\|^2$.
            \State Accumulate $\mathcal{L}_{\rm student}^{\rm pcm}\mathrel{+}=w_{\rm AMD}\ell_{\rm pcm}^{k,j}+\beta_{\rm ema}\ell_{\rm ema}^{k,j}$.
        \EndFor
    \EndFor

    \Statex \textbf{Phase 4: Independent Optimization and EMA Update}
    \State Update only $\theta$ by minimizing $\mathcal{L}_{\rm student}^{\rm pcm}$ over all phase-local pairs.
    \State $\bar\phi\leftarrow\rho_T\bar\phi+(1-\rho_T)\phi$; $\bar\theta\leftarrow\rho_S\bar\theta+(1-\rho_S)\theta$.
\Statex \hspace{-\algorithmicindent}{\textbf{end while}}
\Ensure EMA student $v_{\bar\theta}$ for $K$-step CFG-free inference.
\end{algorithmic}
\end{algorithm}
\paragraph{Shared rollout and decoupled optimization.}
Both base objectives reuse stored rollout states without additional image sampling or reward-model evaluation beyond the teacher RL pipeline. The trajectory and terminal rewards are collected once by the teacher and reused by both optimizers. Although the teacher is updated before the student in each training iteration, the student loss is evaluated on the stored pre-update trajectory and never backpropagates into the teacher. REST therefore preserves the teacher optimizer exactly and changes only how the student consumes its reward-scored rollout.

\paragraph{Difference between the two student objectives.}
The default variant uses one finite-difference velocity target for each student phase. It directly teaches the student to traverse the complete phase in one step and requires $K$ student targets per trajectory. PCM+AMD instead enumerates every adjacent teacher pair inside each phase. The online and EMA students start from different noise levels but are constrained to predict the same phase endpoint. If the selected student boundaries cover the complete $M$-step teacher trajectory, the PCM variant uses $\sum_k(q_{k+1}-q_k)=M$ consistency pairs rather than $K$ segment targets. This explains its higher forward/backward cost.

\paragraph{Role of AMD and EMA.}
In both variants, AMD multiplies the per-sample base distillation loss; it does not modify the segment velocity or PCM endpoint target. Consequently, positive $A_{\rm mix}+b$ strengthens imitation or consistency on preferred trajectories, whereas a negative coefficient reverses the corresponding gradient and discourages low-reward trajectories. The shift $b$ supplies a positive imitation prior during early training. The EMA student plays two related but distinct roles: in the default variant it is an explicit stability regularizer, while in PCM+AMD it additionally provides the stop-gradient phase-consistency target. For both base objectives, gradients update the online student while the student EMA remains a detached target; the four-step DMD extension additionally trains a separate fake-score adapter.

\paragraph{Auxiliary DMD update for four-step students.}
The four-step variant supplements Algorithm~\ref{alg:rest_base} with the auxiliary DMD term in Sec.~\ref{subsec:understanding_amd}, while retaining the EMA student for inference. In the default REST-RAM configuration, each sample independently selects one of the four teacher states aligned to the student input times. The source latent is detached, and the student predicts the clean latent directly in one step. Fake-score and student DMD updates draw their source indices separately; neither evaluates all four source states for each sample. A separate fake-score LoRA is trained by flow matching on re-noised, detached student predictions. For the student update, the EMA teacher and fake-score predictions are evaluated without gradients, and their normalized DMD direction is backpropagated only through the student's one-step prediction. AMD and unweighted DMD contribute to one student optimizer update with coefficients $1$ and $2$, respectively. Fake-score updates do not modify the teacher or student adapters, and no additional images are scored by the reward model.

For the student-rollout source variant, the initial noise and four aligned times are shared with the teacher trajectory. The input state at the randomly selected time is obtained by running the student through the preceding intervals without gradients. Only the final one-step clean-latent prediction receives the DMD gradient; AMD still uses the teacher trajectory.

\begin{figure}[t]
  \centering
  \includegraphics[width=1\linewidth]{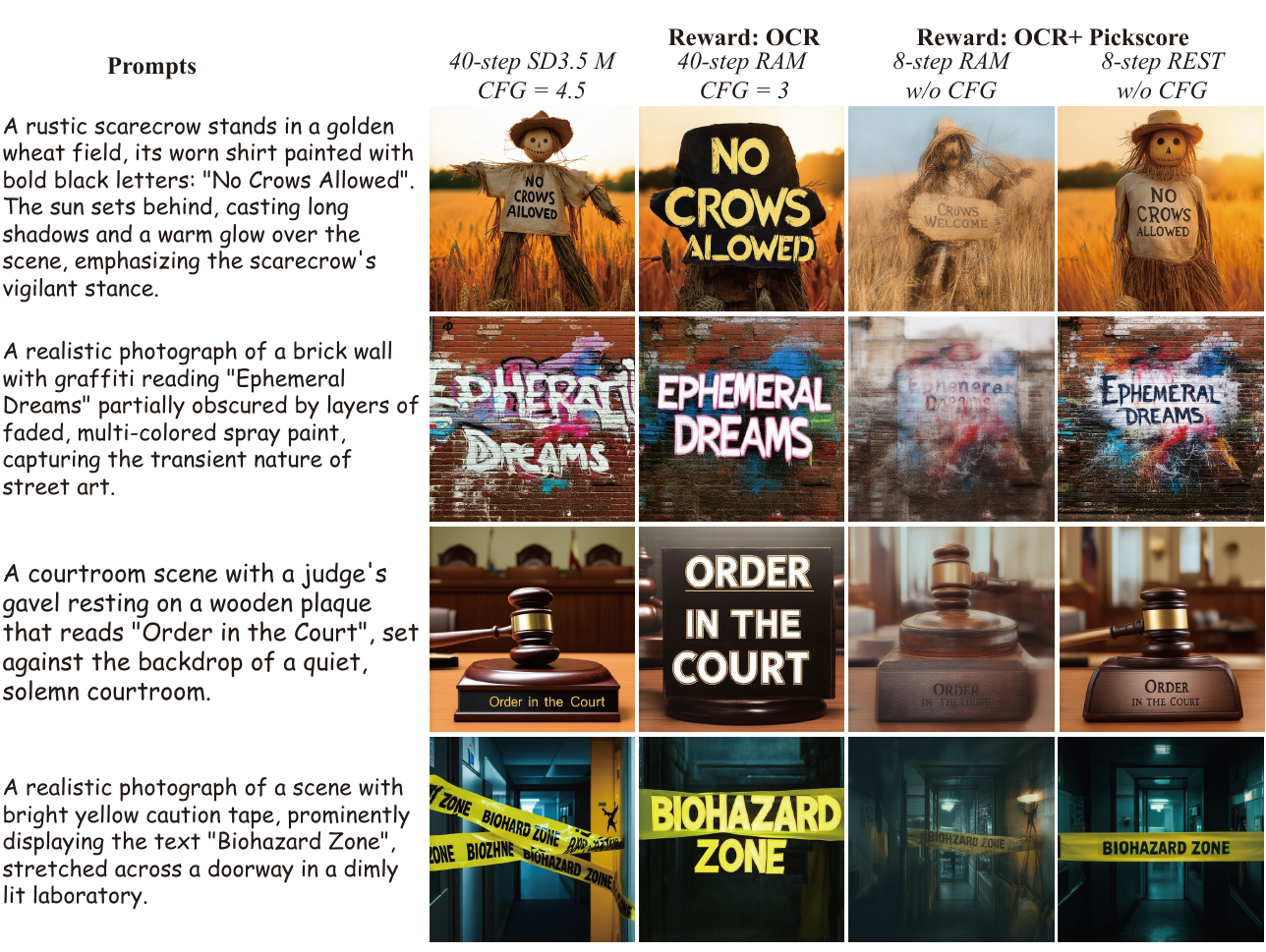}
  \caption{Additional qualitative results on visual text rendering.}
  \label{fig:supp_ocr}
\end{figure}

\begin{figure}[t]
  \centering
  \includegraphics[width=1\linewidth]{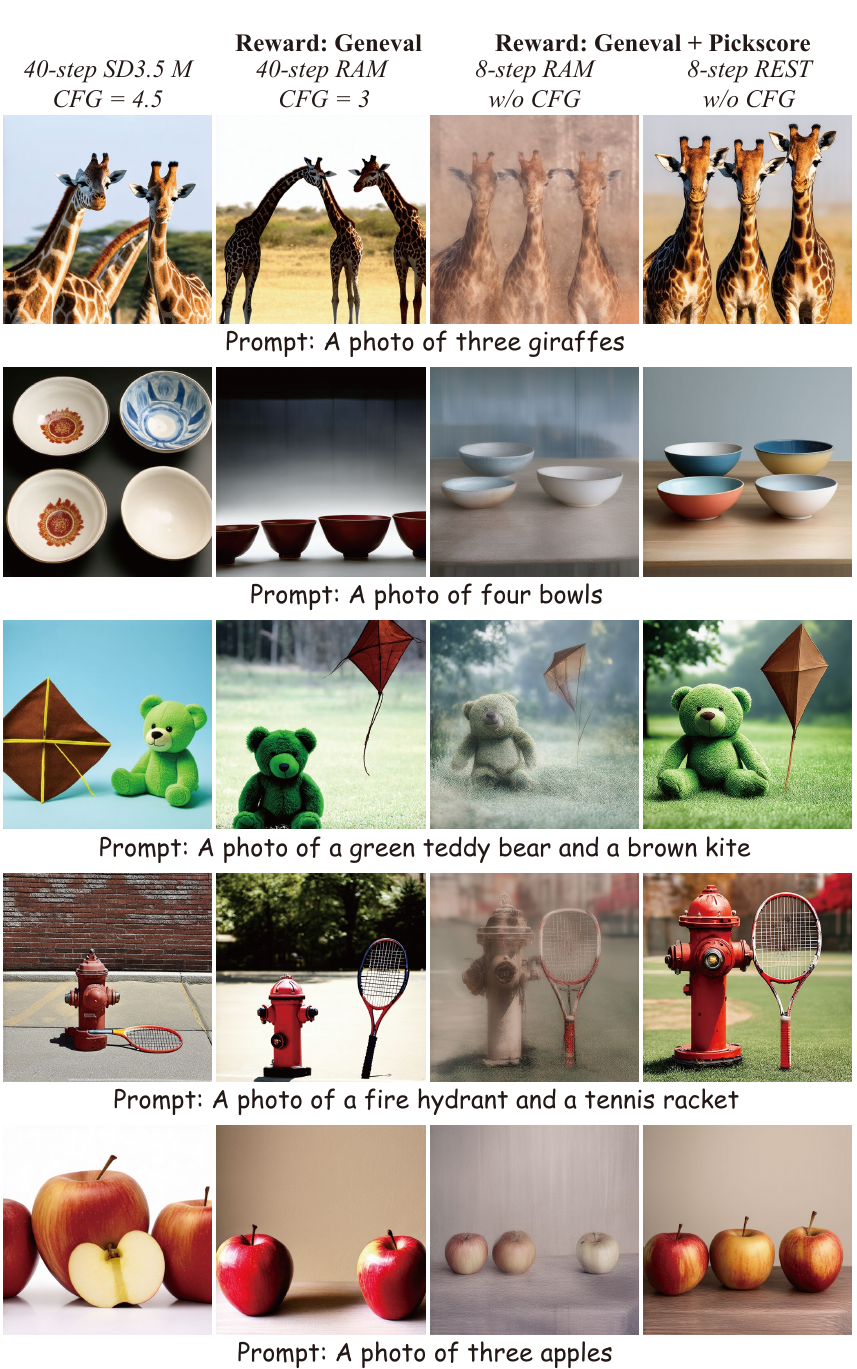}
  \caption{Additional qualitative results on compositional generation (GenEval task).}
  \label{fig:supp_geneval}
\end{figure}

Additional qualitative results for visual text rendering and compositional generation are shown in Figs.~\ref{fig:supp_ocr} and~\ref{fig:supp_geneval}, respectively.

\end{document}